\documentclass[final,5p,times,authoryear]{elsarticle}
\usepackage{natbib}
\usepackage{subcaption}
\usepackage{etoolbox}
\usepackage{tabularx,array}
\newcolumntype{Y}{>{\raggedright\arraybackslash}X}
\AtBeginEnvironment{table*}{%
	\small
	\setlength{\tabcolsep}{5pt}%
}

\AtBeginEnvironment{table}{%
	\small
	\setlength{\tabcolsep}{5pt}% 鍒楅棿璺濈粺涓€
}

\biboptions{authoryear,round}

\makeatletter
\AtBeginDocument{\providecommand{\NAT@and}{\&}\renewcommand{\NAT@and}{\&}}
\makeatother

\usepackage{amsmath,amssymb,graphicx,microtype}
\usepackage{booktabs,tabularx,multirow,adjustbox}
\usepackage{threeparttable}
\usepackage{enumitem}
\setlist[itemize]{leftmargin=*,label=\textbullet,labelsep=0.6em,itemsep=0.25em,topsep=0.3em}

\usepackage{siunitx}
\newcommand{\tablebodyfont}{\footnotesize}

\usepackage[dvipsnames]{xcolor}
\definecolor{CiteSoft}{HTML}{2D9CDB}
\definecolor{URLSoft}{HTML}{2D9CDB}
\usepackage[colorlinks=true,linkcolor=CiteSoft,citecolor=CiteSoft,urlcolor=URLSoft]{hyperref}

\newcommand{\figref}[1]{\hyperref[#1]{Fig.~\ref*{#1}}}
\newcommand{\tabref}[1]{\hyperref[#1]{Table~\ref*{#1}}}

\makeatletter
\AtBeginDocument{%
	\definecolor{CiteSoft}{HTML}{2D9CDB}
	\definecolor{URLSoft}{HTML}{2D9CDB}
	
	\def\@linkcolor{CiteSoft}
	\def\@citecolor{CiteSoft}
	\def\@urlcolor{URLSoft}
	
	\hypersetup{
		colorlinks=true,
		linkcolor=\@linkcolor,
		citecolor=\@citecolor,
		urlcolor=\@urlcolor
	}
}
\makeatother

\journal{Expert Systems with Applications}

\begin{document}

\begin{frontmatter}



\title{DUPLE: An Intelligent Cross-Deployment Recognition Framework for Fiber-Optic Perimeter Security under Scarce Target Labels}


\author[1]{Yifan He}
\ead{yfhe25@m.fudan.edu.cn}

\author[2]{Haodong Zhang}
\ead{zhang_haodong@mail.nwpu.edu.cn}
\author[3]{Qiuheng Song}
\ead{qhsong@fudan.edu.cn}

\author[1]{Lin Lei}
\ead{leilin0603@163.com}

\author[4]{Zhenxuan Zeng}
\ead{zengzhenxuan@mail.nwpu.edu.cn}

\author[1]{Haoyang He}
\ead{hyhe25@m.fudan.edu.cn}

\author[1]{Hongyan Wu\corref{cor1}}
\ead{hywu@fudan.edu.cn}

%% 閫氳浣滆€呰剼娉?\cortext[cor1]{Corresponding author. E-mail: hywu@fudan.edu.cn}

%% 鍗曚綅淇℃伅
\address[1]{College of Smart Materials and Future Energy, 
	Fiber-Optic Research Center, Fudan University, Shanghai 200433, China}

\address[2]{School of Software, Northwestern Polytechnical University, Xi'an 710129, China}

\address[3]{Sichuan Fujinan Technology Co., Ltd., Chengdu 611400, China}

\address[4]{School of Computer Science, Northwestern Polytechnical University, Xi'an 710129, China}


%% Abstract
\begin{abstract}
Distributed Fiber Optic Sensing (DFOS) has emerged as a promising technology for long-range and real-time perimeter security in critical infrastructure monitoring. However, DFOS signals collected from different field deployments often exhibit substantial distribution shifts caused by variations in fiber installation, structural coupling, and environmental noise. These deployment-dependent changes make reliable event recognition difficult in practical perimeter security systems, especially when labeled samples from new target sites are scarce or unavailable.

To address these challenges, this paper proposes DUPLE, an intelligent cross-deployment recognition framework for fiber-optic perimeter security under label-scarce target deployments. DUPLE employs statistically guided meta-learning to enhance recognition robustness across unseen deployments. Specifically, a dual-domain multi-prototype learner jointly models temporal and frequency-domain evidence to capture intra-class variability under deployment shifts. A statistical guidance network estimates sample-specific domain reliability from raw signal statistics, while a query-aware aggregation mechanism adaptively selects relevant prototypes for each test sample.

Extensive experiments on two real-world cross-deployment DFOS benchmarks demonstrate that DUPLE consistently outperforms representative traditional machine learning, deep learning, domain generalization, and meta-learning baselines. Ablation, few-shot, per-deployment, and efficiency analyses further verify the effectiveness and practicality of DUPLE for reliable DFOS-based perimeter security monitoring.

\end{abstract}


%%\item Research highlight 1
%%\item Research highlight 2

%% Keywords
\begin{keyword}
Distributed Fiber Optic Sensing \sep
Perimeter Security \sep
Cross-Deployment Recognition \sep
Label-Scarce Learning \sep
Domain Generalization \sep
Meta-Learning


\end{keyword}

\end{frontmatter}



\section{Introduction}

Distributed Fiber Optic Sensing (DFOS) has emerged as a promising technology for long-range and real-time perimeter security in critical infrastructure monitoring. By interrogating optical-field variations along deployed fiber links, DFOS can acquire continuous vibration measurements over large areas with high temporal resolution and without distributed active electronics along the monitored route~\citep{rao2021recent,hartog2017introduction,wang2020recent}. Compared with conventional cameras, radars, and point sensors, fiber-optic sensing can provide covert, continuous, and privacy-preserving coverage for pipelines, transportation corridors, industrial facilities, and other security-sensitive infrastructures~\citep{munoz2022enhancing,ajo2019distributed,lindsey2019illuminating,tejedor2019contextual,zhong2025intelligent}.

Despite these advantages, reliable event recognition remains difficult when a DFOS monitoring system is transferred from one field deployment to another. Many existing recognition models are developed and evaluated under fixed-site or randomly split settings, where training and testing samples share similar installation conditions. In practical perimeter security systems, however, a trained model may be deployed on a new fence, wall, cable layout, or coupling structure with different ambient noise and mechanical transfer characteristics. Such deployment-dependent changes alter both temporal waveforms and spectral patterns, causing severe distribution shifts~\citep{munoz2022enhancing}. Moreover, labeled samples from new target deployments are often scarce and expensive to obtain during early deployment. As a result, conventional supervised retraining is often impractical, and models that perform well in a source deployment may degrade sharply when transferred to an unseen target deployment.

Existing DFOS recognition methods only partially address this problem. Traditional approaches extract hand-crafted time- and frequency-domain descriptors and classify events using support vector machines, random forests, hidden Markov models, or related classifiers~\citep{he2021optical,tejedor2019contextual}. Recent deep learning methods, including one- or two-dimensional CNNs and recurrent architectures, learn discriminative representations directly from raw or time-frequency signals~\citep{wu2019one,wu2021pattern}. Although these methods can achieve strong performance under fixed sensing conditions, they usually assume that training and testing data follow similar distributions. Domain adaptation and transfer learning can improve robustness when target data are available for fine-tuning or distribution alignment, but this assumption is often unrealistic in label-scarce perimeter deployments. Domain generalization and meta-learning are therefore more suitable candidates: by learning across multiple source deployments and episodic support/query splits, a model can acquire adaptation behavior that is better aligned with unseen-site recognition~\citep{finn2017model,snell2017prototypical,ye2020few,luong2023few}.

However, applying meta-learning to DFOS perimeter security still faces three key challenges. First, DFOS events are naturally multi-modal across deployments. Time-domain and frequency-domain evidence may carry complementary information, and a single prototype per class is often insufficient to represent diverse event patterns induced by different installation structures and coupling conditions~\citep{zhang2022varphi,allen2019infinite}. Second, the reliability of each signal view varies from sample to sample. Raw statistical properties such as energy concentration, spectral dispersion, and noise-related descriptors provide useful cues, but they are rarely used to guide cross-domain prototype construction and fusion~\citep{titov2022quantification}. Third, most metric-based meta-classifiers compare each query with fixed class prototypes. For interferometric fiber signals affected by local artifacts and deployment-specific distortions, the decision process should adaptively select the most relevant class evidence for each query rather than relying on a static class summary~\citep{ye2020few,vinyals2016matching}.

To address these challenges, this paper proposes DUPLE, an intelligent cross-deployment recognition framework for fiber-optic perimeter security under label-scarce target deployments. In practical security operation, DUPLE serves as a deployment-aware recognition module that converts raw vibration measurements into robust class evidence, estimates the reliability of complementary signal views from statistical descriptors, and produces adaptive decisions without extensive target-domain retraining. The framework is built on statistically guided meta-learning and integrates dual-domain multi-prototype representation, statistical reliability guidance, and query-aware prototype aggregation.

The main contributions of this work are summarized as follows:

\begin{itemize}
	
	\item We formulate and evaluate a label-scarce cross-deployment recognition problem for DFOS-based perimeter security, where a model trained on existing fiber installations must generalize to unseen target deployments with limited labeled samples. This setting better reflects practical perimeter monitoring than conventional fixed-site or random-split evaluation.
	
	\item We propose DUPLE, an intelligent cross-deployment recognition framework based on statistically guided meta-learning. DUPLE combines dual-domain multi-prototype representation, statistical reliability guidance, and query-aware prototype aggregation to construct deployment-tolerant class evidence from limited support samples.
	
	\item We evaluate DUPLE on two real-world cross-deployment DFOS benchmarks under a Leave-One-Deployment-Out protocol. Comparisons with traditional machine learning, deep learning, domain-generalization, and meta-learning baselines, together with ablation studies, few-shot analysis, per-deployment evaluation, per-class analysis, and efficiency assessment, verify its robustness and practicality for fiber-optic perimeter security monitoring.
\end{itemize}

The remainder of this paper is organized as follows.
\hyperref[sec:related_work]{Section~\ref*{sec:related_work}} reviews related work on DFOS event recognition, domain generalization, and meta-learning.
\hyperref[sec:datasets]{Section~\ref*{sec:datasets}} introduces the cross-deployment datasets and evaluation setting.
\hyperref[sec:method]{Section~\ref*{sec:method}} presents the proposed DUPLE framework.
\hyperref[sec:experiments]{Section~\ref*{sec:experiments}} reports experimental results and analysis, and \hyperref[sec:conclusion]{Section~\ref*{sec:conclusion}} concludes the paper.

\section{Related Work}
\label{sec:related_work}

\subsection{DFOS-based perimeter event recognition}

Distributed Fiber Optic Sensing (DFOS) has been widely investigated for vibration-based event recognition in perimeter security and critical infrastructure monitoring. Early studies usually followed a feature-engineering pipeline, where time-domain, frequency-domain, or time-frequency descriptors were extracted from fiber-optic vibration signals and then classified by conventional machine learning models. Short-time Fourier transform and related time-frequency descriptors have been used to characterize intrusion-induced vibration patterns~\citep{wu2022improved,madsen2007intruder}. Wavelet coefficients and Mel-frequency cepstral coefficients have also been explored to represent transient and spectral characteristics of fiber-optic signals~\citep{jia2019event,wu2021pattern}. Based on these descriptors, support vector machines, Gaussian mixture models, hidden Markov models, and related classifiers have been widely used for event recognition~\citep{qu2010svm,xu2017pattern,tejedor2019contextual,tejedor2017novel,wu2019dynamic}. These methods are interpretable and computationally efficient, but their performance strongly depends on manually designed features and stable sensing conditions. When the installation structure, coupling state, background noise, or monitored object changes, the extracted descriptors may no longer preserve the same class-discriminative patterns.

Deep learning methods have reduced the need for hand-crafted feature design by learning representations directly from raw signals, time-frequency maps, or multi-channel sensing measurements. Recurrent neural networks and long short-term memory networks have been introduced to capture temporal dependencies in sequential vibration signals~\citep{medsker2001recurrent,sherstinsky2020fundamentals}. CNN-based and CNN-RNN hybrid architectures further combine local pattern extraction with temporal modeling for fiber-optic event recognition~\citep{wang2016cnn,khaki2020cnn,wu2019one,wu2021pattern}. Attention-based and Transformer-style models provide additional flexibility for modeling long-range contextual information and multi-scale temporal structures~\citep{vaswani2017attention,lim2021temporal,zhou2021informer}. More recent studies have further improved DFOS-based perimeter event recognition by designing domain-informed neural architectures. For example, M2former integrates multi-scale feature extraction, multi-graph attention embedding, and event-specific attention for single- and multi-event recognition~\citep{lin2025m2former}. Bai et al. developed a dual-branch spatiotemporal synergistic network for critical infrastructure perimeter security, where temporal and spatial signal patterns are jointly modeled with physics-inspired data augmentation and noise-aware learning~\citep{bai2026dual}. These studies demonstrate that physically informed feature modeling and specialized spatiotemporal architectures can substantially improve supervised DFOS recognition performance.

Nevertheless, most existing DFOS recognition studies, including recent deep models, are mainly evaluated under fixed-site, supervised, or randomly split protocols. In such settings, training and test samples usually share similar installation conditions and environmental characteristics. This assumption is often inconsistent with practical perimeter security deployment, where a model trained on existing fiber installations may need to recognize events at newly deployed sites with different fence structures, cable layouts, coupling conditions, and ambient noise. Therefore, while existing studies have advanced supervised DFOS event recognition, label-scarce cross-deployment recognition remains insufficiently explored.

\subsection{Domain generalization for deployment shift}

Deployment-induced distribution shift is a major obstacle to reliable DFOS-based perimeter event recognition. In practical sensing systems, changes in installation method, coupling condition, monitored structure, and environmental noise can significantly affect the temporal and spectral characteristics of vibration signals~\citep{munoz2022enhancing}. As a result, the identical-distribution assumption between training and test data is often violated when a model is transferred from one fiber installation to another. This problem is particularly challenging in perimeter security applications, where collecting sufficient labeled samples from every new deployment can be time-consuming and expensive.

Domain adaptation and domain generalization provide two related ways to improve robustness under distribution shift. Domain adaptation methods usually reduce the discrepancy between source and target distributions by using target-domain data during training or adaptation. However, this requirement can be restrictive for perimeter security systems, especially during early deployment stages when target labels are scarce, expensive, or unavailable. Domain generalization instead aims to learn models that can generalize to unseen target domains using only source-domain data. This setting is more consistent with cross-deployment DFOS recognition, because multiple existing fiber installations can be treated as source domains during training, while a new deployment is left unseen until evaluation.

However, conventional domain generalization methods usually learn a static domain-invariant model and do not explicitly exploit limited support samples from a new target deployment. In practical DFOS-based perimeter security, a small number of labeled samples may be obtainable during early site commissioning, and such samples can provide valuable deployment-specific information. Moreover, generic domain generalization objectives are not designed for the signal-specific characteristics of DFOS data, such as the complementary roles of temporal and spectral evidence, deployment-dependent noise patterns, and class-wise intra-deployment variability. These limitations motivate a recognition framework that can combine cross-domain generalization with label-scarce target adaptation.

\subsection{Meta-learning and prototype-based recognition}

Meta-learning provides a promising direction for label-scarce cross-deployment recognition. Instead of training a single classifier only for the source distribution, meta-learning constructs episodic tasks during training and encourages the model to acquire adaptation behavior that transfers to new tasks or domains. Model-Agnostic Meta-Learning learns an initialization that can be rapidly fine-tuned with a few labeled samples~\citep{finn2017model}, while metric-based methods such as Matching Networks and Prototypical Networks classify query samples according to distances in an embedding space~\citep{vinyals2016matching,snell2017prototypical}. These methods have achieved strong performance in few-shot learning and have inspired many extensions, including ridge-regression-based classifiers, representation adaptation modules, and transformer-based support-set adaptation~\citep{bertinetto2018meta,raghu2019rapid,ye2020few,luong2023few}.

\begin{figure*}[htbp]
	\centering
	\includegraphics[width=0.85\textwidth]{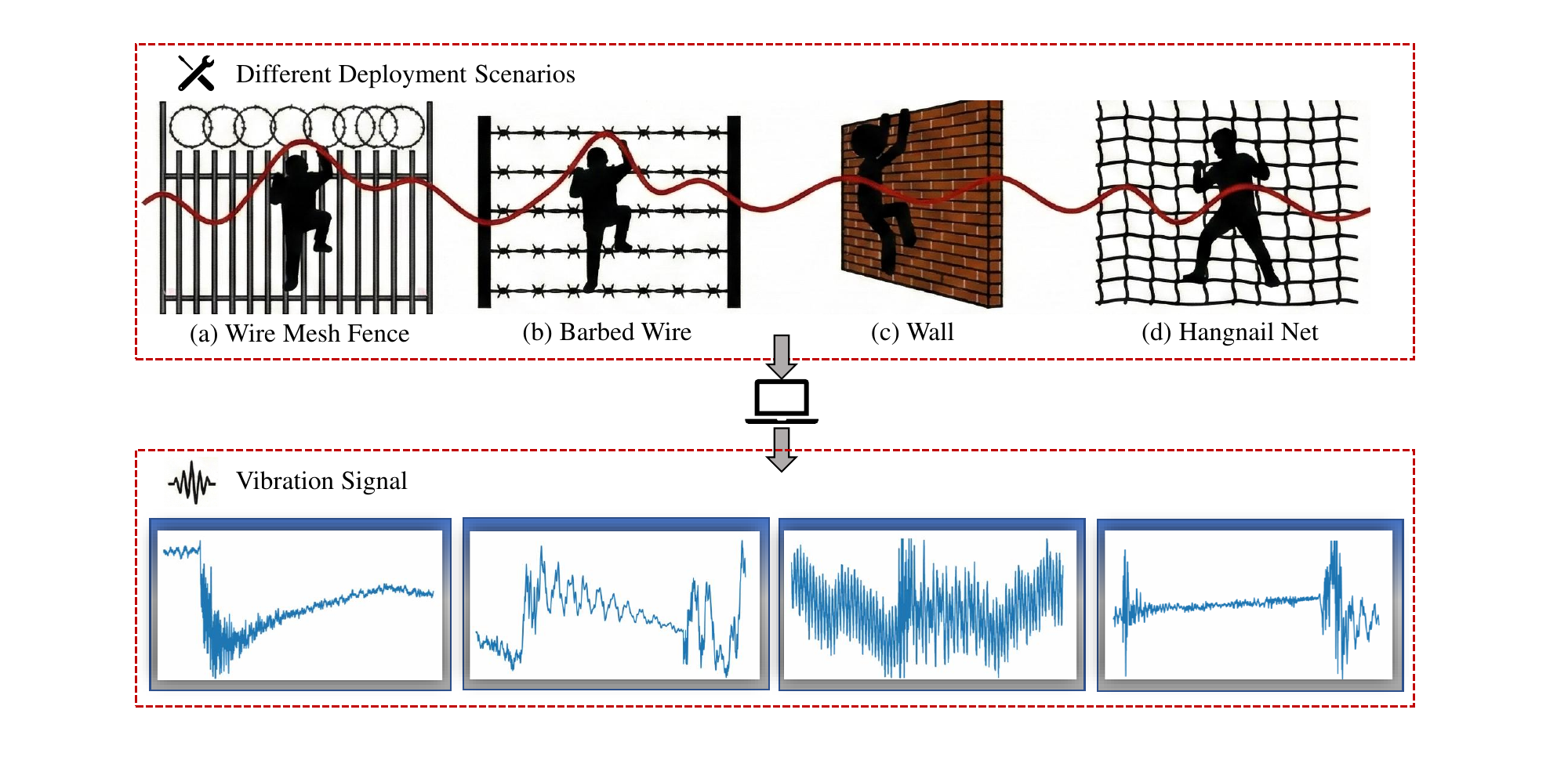}
	\caption{Climbing vibration signals collected under different deployment configurations. (a) Wire mesh fence. (b) Barbed wire. (c) Wall. (d) Hangnail net.}
	\label{fig:signal}
\end{figure*}

Despite these advances, directly applying standard meta-learning methods to DFOS perimeter security remains limited. First, many metric-based methods represent each class with a single prototype, which is insufficient when the same event category exhibits multiple waveform or spectral patterns across different fiber installations~\citep{snell2017prototypical,zhang2022varphi,allen2019infinite}. Second, most few-shot learners treat all feature views or support samples with fixed reliability, whereas DFOS signals often contain deployment-specific noise and view-dependent degradation~\citep{titov2022quantification}. Third, conventional prototype classifiers usually compare each query with static class summaries, although practical fiber-optic signals may require query-specific evidence selection under local artifacts and changing coupling conditions~\citep{vinyals2016matching,ye2020few}. These limitations suggest that prototype representations and aggregation mechanisms should be aware of both signal statistics and deployment-induced variability.

In summary, existing DFOS event recognition studies have achieved substantial progress under supervised and fixed-site evaluation settings, and recent domain-informed and spatiotemporal neural architectures further show the value of physically informed feature modeling for perimeter security. However, label-scarce cross-deployment recognition remains insufficiently explored. Existing domain generalization methods usually learn static domain-invariant models and do not explicitly exploit limited target support samples, while standard meta-learning methods are not designed to handle the multi-modal class distributions, sample-dependent view reliability, and query-specific evidence selection required by DFOS signals under deployment shifts. To fill this gap, this paper proposes DUPLE, a statistically guided dual-domain meta-learning framework that integrates multi-prototype representation, statistical reliability guidance, and query-aware aggregation for robust fiber-optic perimeter security across unseen deployments.
                                                
%% Use \section commands to start a sectiont
\section{Cross-Deployment Datasets and Evaluation Setting}
\label{sec:datasets}

This section describes the DFOS datasets and evaluation protocol used to assess cross-deployment recognition under scarce target labels. Each fiber installation configuration is treated as a deployment domain. A model is trained on source deployment(s) and evaluated on a held-out target deployment, where only a small number of labeled samples may be available as support data.

In our DFOS system, event signals are obtained by phase-demodulating the output of a Michelson interferometer. \figref{fig:signal} compares climbing signals collected from different deployment configurations. Although these signals correspond to the same event type, they exhibit clear differences in amplitude, frequency distribution, and pulse characteristics due to variations in installation location, contact material, distance from the vibration source, and coupling condition. These differences illustrate the deployment-induced domain shift considered in this study.

We construct two real-world DFOS benchmarks, OSDG1 and OSDG2. Each benchmark contains multiple deployment domains with a consistent label space within that benchmark, enabling controlled cross-deployment evaluation. OSDG1 focuses on fence-based perimeter deployments with four event categories, whereas OSDG2 covers more diverse structural materials with three event categories. The two benchmarks differ in deployment structure, class composition, and class imbalance, providing complementary test cases for DFOS recognition under deployment shift. All signals are preprocessed to a unified length before being fed into the model.

The OSDG1 dataset contains two deployment scenarios. In the first scenario, fiber optic cables are installed on a traditional barbed wire fence. In the second scenario, fiber optic cables are fixed to a composite boundary structure consisting of a lower railing and an upper spiral barbed wire fence. Both scenarios record four types of events: background, human climbing, rain, and smash/impact events. \tabref{tab:OSDG1} reports the sample distribution of OSDG1. The class imbalance reflects the difficulty of collecting specific event samples in real-world perimeter security projects.

\begin{table}[htbp]
	\centering
	\begin{threeparttable}
	\caption{Sample distribution of the OSDG1 benchmark.}
	\label{tab:OSDG1}
	\begingroup
	\tablebodyfont
	\begin{tabular}{@{}lllll@{}}
		\toprule
		Deployment method & Background & Climb & Rain & Smash \\
		\midrule
		Barbed wire       & 70 & 125 & 80 & 104 \\
		Wire mesh fence   & 125 & 43 & 37 & 123 \\
		\bottomrule
	\end{tabular}
	\endgroup
	\end{threeparttable}
\end{table}

The OSDG2 dataset contains three deployment conditions: hangnail net, solid concrete wall, and railing. Each deployment includes three event categories: background, ladder climbing, and impact. \tabref{tab:OSDG2} reports the sample distribution of OSDG2. Compared with OSDG1, OSDG2 emphasizes ladder-assisted climbing and more diverse impact sources, including tool strikes, rock impacts, and wildlife impacts.

\begin{table}[htbp]
	\centering
	\begin{threeparttable}
	\caption{Sample distribution of the OSDG2 benchmark.}
	\label{tab:OSDG2}
	\begingroup
	\tablebodyfont
	\begin{tabular}{@{}llll@{}}
		\toprule
		Deployment method & Background & Ladder climbing & Impact \\
		\midrule
		Hangnail net & 200 & 130 & 199 \\
		Wall         & 200 & 133 & 200 \\
		Railing      & 200 & 131 & 200 \\
		\bottomrule
	\end{tabular}
	\endgroup
	\end{threeparttable}
\end{table}

\begin{figure*}[htbp]
	\centering
	\includegraphics[width=0.99\textwidth]{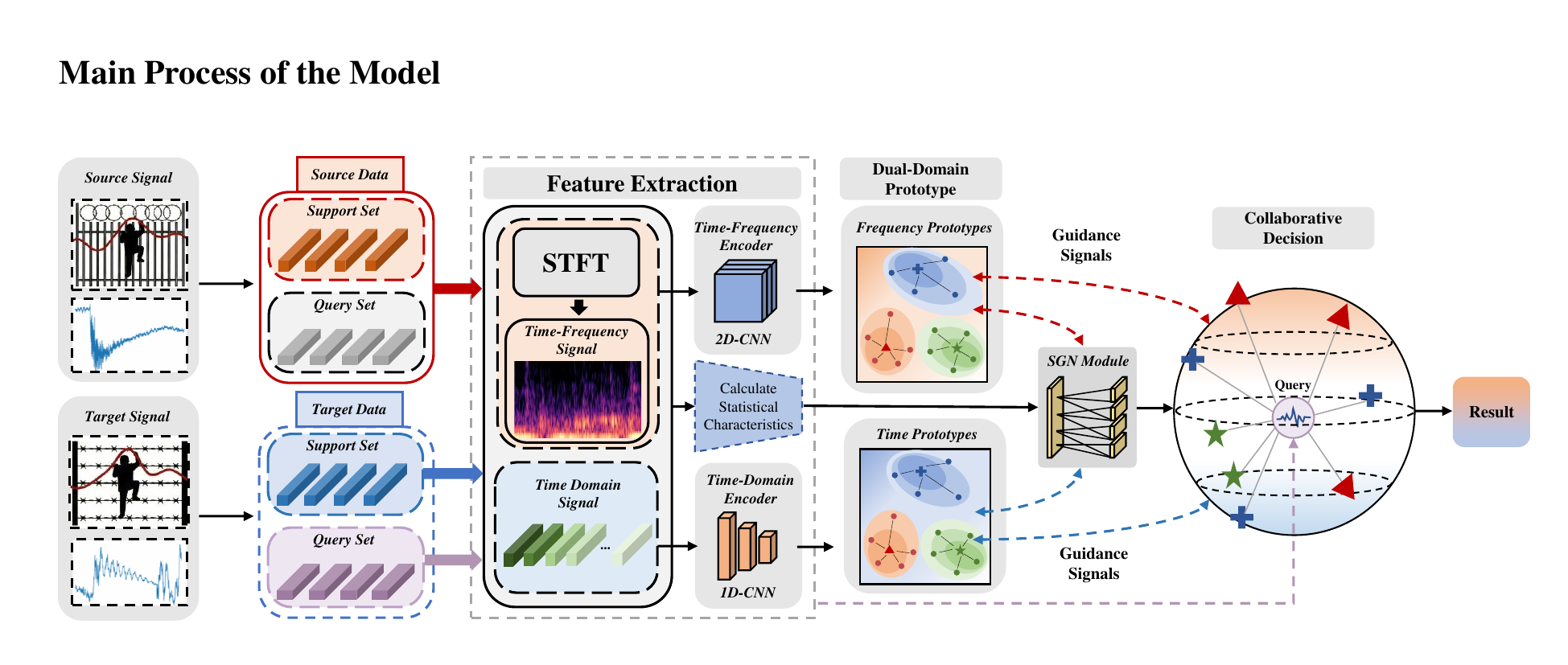}
	\caption{Overview of the proposed DUPLE framework.}
	\label{fig:main_model}
\end{figure*}

We adopt a Leave-One-Deployment-Out (LODO) protocol. In each round, one deployment domain is held out as the target deployment, and the remaining deployment domain(s) are used as source deployment(s). Under label-scarce target conditions, only $K$ labeled samples per class from the target deployment are used as support samples, while the remaining target samples are used as query samples for evaluation. Query samples are not used for model training or parameter updating unless otherwise specified.

\section{Method}
\label{sec:method}

\subsection{Overall Framework Overview}

DUPLE is designed as a label-scarce cross-deployment recognition framework for DFOS-based perimeter security. Under the episodic meta-learning setting, each task contains a small labeled support set and a query set. The goal is to construct reliable class evidence from limited support samples and predict query labels under deployment-induced distribution shifts.

As illustrated in \figref{fig:main_model}, DUPLE consists of three coordinated components: a dual-domain multi-prototype meta-learner, a Statistical Guidance Network (SGN), and a collaborative decision module. Each vibration signal is processed through two parallel branches: a time-domain branch that takes the original waveform as input and a time-frequency branch that takes the STFT spectrogram as input. In parallel, a 26-dimensional statistical descriptor is computed from the raw signal and fed into the SGN to estimate sample-specific reliability cues.

For each episode, the dual-domain meta-learner constructs multiple class prototypes in both branches, allowing each event category to be represented by heterogeneous class evidence. The collaborative decision module then integrates dual-branch predictions through SGN-guided reliability weighting and query-aware prototype aggregation. All components are trained jointly in an episodic manner, enabling DUPLE to recognize events in a new deployment using a small support set without additional parameter fine-tuning.

\subsection{Dual-Domain Multi-Prototype Meta-Learner}

This module addresses intra-class heterogeneity under deployment shifts by representing each event class with multiple prototypes in both the time-domain and time-frequency branches. Each raw vibration signal is processed as a waveform input $x^{(t)}$ and an STFT spectrogram input $x^{(f)}$.

In our implementation, the frequency-domain input $x^{(f)}$ is constructed by Short-Time Fourier Transform (STFT). 
Given a raw 1-D signal segment sampled at $f_s=50\,\mathrm{kHz}$, we compute STFT using a Hann window with FFT size $n_{\mathrm{fft}}=1024$ and hop length $h=512$ (i.e., 50\% overlap). 
Let $Z \in \mathbb{C}^{F \times T}$ denote the complex STFT; we take the magnitude spectrogram and apply logarithmic compression: 
$x^{(f)}=\log(|Z|+\epsilon)$ with $\epsilon=10^{-8}$. 
We then apply per-spectrogram standardization to zero mean and unit variance to stabilize the dynamic range. 
Finally, the spectrogram is padded/cropped to a fixed resolution of $1 \times 513 \times 294$ (channel $\times$ frequency bins $\times$ time frames) before feeding into the frequency encoder.

We use two parallel feature extractors, $f_t(\cdot)$ and $f_f(\cdot)$, where the 1DCNN extracts time-domain signal features and the 2DCNN extracts signal time-spectrogram features. They encode $x^{(t)}$ and $x^{(f)}$ into $D$-dimensional feature vectors, $\mathbf{z}^{(t)} = f_t(x^{(t)}) \in \mathbb{R}^D$ and $\mathbf{z}^{(f)} = f_f(x^{(f)}) \in \mathbb{R}^D$. These embedding networks are designed to produce comparable feature representations for both domains (of dimension $D=128$ in our implementation). Given a labeled support set, we obtain a set of support embeddings in the time domain ${\mathbf{z}{i}^{(t)}}$ and the frequency domain ${\mathbf{z}{i}^{(f)}}$. These two domains are handled independently by the same meta-learning logic, as described below.

\begin{figure}[htbp]
	\centering
	\includegraphics[width=0.8\linewidth]{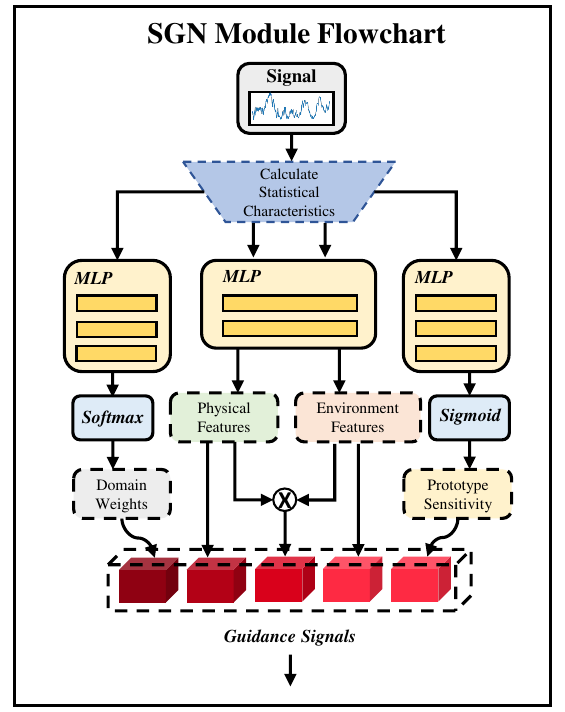}
	\caption{Structure of the Statistical Guidance Network (SGN).}
	\label{fig:sgn}
\end{figure}

We introduce an adaptive multi-prototype representation to capture the complex class distribution, rather than collapsing the support examples for each class into a single prototype as in standard prototypical networks. For each class $c$, we cluster the embeddings of the support examples ${\mathbf{z}{i}^{(t)}: y_i = c}$ and ${\mathbf{z}{i}^{(f)}: y_i = c}$, generating $K_c^{(t)}$ and $K_c^{(f)}$ prototypes in the time and frequency domains, respectively. The number of prototypes $K_c$ is dynamically determined based on the characteristics of the support set. This strategy is particularly useful when dealing with classes that exhibit diverse patterns, as it allows the model to represent the class with multiple prototype vectors, thereby enhancing its generalization ability.

For each class $c$ in each domain $d \in \{t,f\}$, we determine the number of prototypes $K_c^{(d)}$ in a data-driven manner.
Let $\mathcal{Z}_c^{(d)}=\{\mathbf{z}_i^{(d)} \mid y_i=c\}$ be the support embeddings of class $c$ and $N_c=|\mathcal{Z}_c^{(d)}|$.
We run $K$-means clustering on $\mathcal{Z}_c^{(d)}$ with candidate $K \in \{1,\ldots,\min(K_{\max}, N_c)\}$ and obtain $K$ centroids as prototypes.
To select $K_c^{(d)}$, we compute the within-cluster sum of squares $\mathrm{WCSS}(K)$ and minimize a penalized criterion:
\begin{equation}
	J(K) = \mathrm{WCSS}(K) + \alpha \cdot K \cdot \log(\max(2, N_c)),
\end{equation}
where $\alpha$ controls the complexity penalty.The selected $K_c^{(d)}=\arg\min_K J(K)$ balances intra-class multi-modality and over fragmentation.In our experiments, we set $K_{\max}=3$ and $\alpha=0.003$.

After obtaining prototypes for each class in both domains, we compute the matching score between each query sample and the class prototype. Specifically, for each query sample $\mathbf{z}_q$, we use cosine similarity to measure the similarity between the query and each prototype. For domain $d \in {t, f}$, we calculate the similarity between the query embedding $\mathbf{z}q^{(d)}$ and each prototype $\mathbf{c}_{c,k}^{(d)}$ of class $c$. The matching score of class $c$ in domain $d$ is calculated as follows:
\begin{equation}
	\ell_{c}^{(d)}(q) = \log \sum_{k=1}^{K_c} \exp\Big(\lambda \cdot \text{sim}(\mathbf{z}_q, \mathbf{c}_{c,k}^{(d)})\Big) \,.
\end{equation}
where $\text{sim}(\cdot, \cdot)$ is the cosine similarity between the query and prototype, and $\lambda$ is a temperature scaling factor that controls the sharpness of the distribution. The result is a logit score $\ell_{c}^{(d)}$ for each class $c$ in both domains. The logits from both domains are then passed to the Collaborative Decision module for further fusion and decision-making.

\subsection{Statistical Guidance Network (SGN)}

The SGN estimates sample-specific reliability cues from raw signal statistics, allowing DUPLE to adjust the contributions of the time-domain and time-frequency branches. Concretely, for each signal we extract a 26-dimensional statistical feature vector $\mathbf{s}$ composed of 16 time-domain features and 10 frequency-domain features . The definitions of these time-domain and frequency-domain features are summarized in \tabref{tab:time_features} and \tabref{tab:freq_features}, respectively. These statistics capture global amplitude, energy and spectral characteristics that are not immediately apparent from the raw time-series or STFT representations, and are used to guide the meta-learner, provide domain-specific weights, and enhance the prototype matching process. The specific structure of the SGN is shown in \figref{fig:sgn}.

\begin{figure}[htbp]
	\centering
	\includegraphics[width=\linewidth]{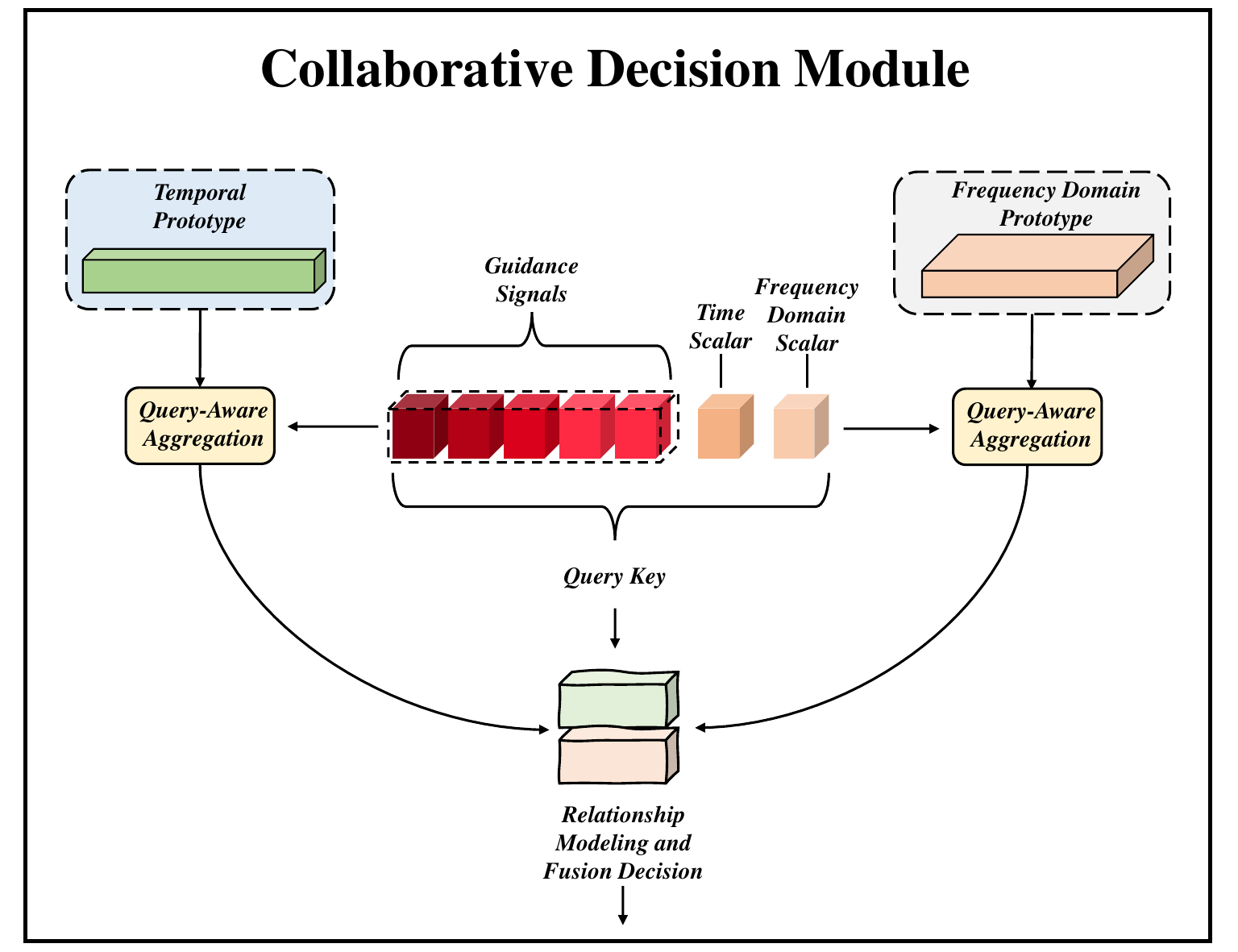}
	\caption{Structure of the collaborative decision module.}
	\label{fig:CDM}
\end{figure}

\begin{table*}[t]
	\centering
	\begin{threeparttable}
	\caption{Time-domain statistical features.}
	\label{tab:time_features}
	\tablebodyfont
	\begin{tabular}{l c l c}
		\hline
		Feature & Formula & Feature & Formula \\
		\hline
		Mean ($S_1$) &
		$\mu = \frac{1}{N}\sum_{n=1}^{N} x_n$
		&
		Std. dev. ($S_2$) &
		$\sigma = \sqrt{\frac{1}{N}\sum_{n=1}^{N}(x_n-\mu)^2}$ \\
		Variance ($S_3$) &
		$\sigma^{2} = \frac{1}{N}\sum_{n=1}^{N}(x_n-\mu)^2$
		&
		Maximum ($S_4$) &
		$\max_{n} x_n$ \\
		Minimum ($S_5$) &
		$\min_{n} x_n$
		&
		Peak-to-peak ($S_6$) &
		$\max_{n} x_n - \min_{n} x_n$ \\
		RMS ($S_7$) &
		$\sqrt{\frac{1}{N}\sum_{n=1}^{N} x_n^{2}}$
		&
		Skewness ($S_8$) &
		$\mathrm{skew}(x) = \frac{1}{N}\sum_{n=1}^{N}
		\big(\frac{x_n-\mu}{\sigma}\big)^{3}$ \\
		Kurtosis ($S_9$) &
		$\mathrm{kurt}(x) = \frac{1}{N}\sum_{n=1}^{N}
		\big(\frac{x_n-\mu}{\sigma}\big)^{4}$
		&
		Mean abs. value ($S_{10}$) &
		$\frac{1}{N}\sum_{n=1}^{N}\lvert x_n\rvert$ \\
		Median ($S_{11}$) &
		$\operatorname{median}(\{x_n\})$
		&
		IQR ($S_{12}$) &
		$Q_{3} - Q_{1}$ \\
		Signal energy ($S_{13}$) &
		$E = \sum_{n=1}^{N} x_n^{2}$
		&
		Average energy ($S_{14}$) &
		$\frac{E}{N}$ \\
		Zero-crossing rate ($S_{15}$) &
		$\mathrm{ZCR} = \frac{N_{\text{zc}}}{N}$
		&
		Peak density ($S_{16}$) &
		$\mathrm{PD} = \frac{N_{\text{peaks}}}{N}$ \\
		\hline
	\end{tabular}
	\end{threeparttable}
\end{table*}

\begin{table*}[t]
	\centering
	\begin{threeparttable}
	\caption{Frequency-domain statistical features.}
	\label{tab:freq_features}
	\tablebodyfont
	\begin{tabular}{l c l c}
		\hline
		Feature & Formula & Feature & Formula \\
		\hline
		Spec. mean mag. ($S_{17}$) &
		$\frac{1}{K}\sum_{k=1}^{K} \lvert X(f_k)\rvert$
		&
		Spec. std. ($S_{18}$) &
		$\sqrt{\frac{1}{K}\sum_{k=1}^{K}
			\big(\lvert X(f_k)\rvert-\bar{A}\big)^{2}}$ \\
		Spec. max mag. ($S_{19}$) &
		$\max_{k}\lvert X(f_k)\rvert$
		&
		Spec. min mag. ($S_{20}$) &
		$\min_{k}\lvert X(f_k)\rvert$ \\
		Spectral centroid ($S_{21}$) &
		$f_c = \dfrac{\sum_{k} f_k \lvert X(f_k)\rvert}
		{\sum_{k} \lvert X(f_k)\rvert}$
		&
		Spectral bandwidth ($S_{22}$) &
		$\sqrt{\dfrac{\sum_{k} (f_k - f_c)^{2} \lvert X(f_k)\rvert}
			{\sum_{k} \lvert X(f_k)\rvert}}$ \\
		Spectral roll-off ($S_{23}$) &
		$\sum_{f_k \le f_r} \lvert X(f_k)\rvert^{2}
		\ge 0.95\sum_{k}\lvert X(f_k)\rvert^{2}$
		&
		Spectral flatness ($S_{24}$) &
		$\dfrac{\exp\big(\frac{1}{K}\sum_{k=1}^{K}
			\ln \lvert X(f_k)\rvert\big)}
		{\frac{1}{K}\sum_{k=1}^{K}\lvert X(f_k)\rvert}$ \\
		Dominant frequency ($S_{25}$) &
		$f_{\text{dom}} = f_{k^\ast},
		\ k^\ast = \arg\max_{k}\lvert X(f_k)\rvert$
		&
		Spectral entropy ($S_{26}$) &
		$H = -\sum_{k} p_k \log_{2} p_k,\ 
		p_k = \dfrac{\lvert X(f_k)\rvert}
		{\sum_{j}\lvert X(f_j)\rvert}$ \\
		\hline
	\end{tabular}
	\end{threeparttable}
\end{table*}

The SGN receives as input the 26-dimensional vector $\mathbf{s}$ for each sample. This vector encodes various global properties of the signal, capturing information beyond the original time-domain or frequency-domain representations. The network consists of two sub-networks: one for extracting physical features and the other for extracting environmental context. Each sub-network processes $\mathbf{s}$ and outputs latent vectors $\mathbf{p}$ and $\mathbf{e}$, representing the physical and environmental characteristics of the signal, respectively.

These latent vectors are concatenated and fed into a guidance signal generation network, which generates a guidance vector $\mathbf{g}$ that informs the meta-learner and decision module. The guidance vector is generated using the tanh activation function:
\begin{equation}
\mathbf{g} = \tanh(W_g [\mathbf{p}; \mathbf{e}] + \mathbf{b}_g) \,.
\end{equation}

where $[\mathbf{p}; \mathbf{e}]$ is the concatenated vector of physical and environmental features, and $W_g$ and $\mathbf{b}_g$ are learnable weights and biases. The SGN also produces domain importance weights $\alpha_t$ and $\alpha_f$ for the time and frequency domains, respectively, which indicate the relative relevance of each domain. These weights are learned through a softmax function over the concatenated feature vector $\mathbf{s}$:
\begin{equation}
	\alpha_t, \alpha_f = \text{softmax}(W_{\alpha} \mathbf{s} + b_{\alpha}) \,.
\end{equation}

In addition, the SGN generates a prototype sensitivity scalar $\beta$, which controls how strict or flexible the prototype matching should be. This scalar is learned to adapt based on the characteristics of the input sample. The outputs from the SGN, including the guidance vector $\mathbf{g}$, domain importance weights $\alpha_t, \alpha_f$, and prototype sensitivity $\beta$, are passed to the Collaborative Decision module, which uses them to adjust the predictions from the time and frequency domains.

\subsection{Collaborative Decision with Query-Aware Attention}
This module performs query-specific evidence selection by aggregating the most relevant prototypes from the time-domain and time-frequency branches. It uses the SGN guidance cues and initial dual-branch predictions to produce the final query decision. Its specific structure is shown in \figref{fig:CDM}. For each query sample, the module computes a query-specific key vector $\mathbf{q}_c$, which is a combination of the query's guidance vector $\mathbf{g}$ and its initial logits from the meta-learner. The query key vector $\mathbf{q}_c$ is used to compute attention weights for each prototype in the time and frequency domains. Specifically, the attention weight for the $k$-th prototype of class $c$ in domain $d$ is calculated as:
\begin{equation}
a_{c,k}^{(d)} = \frac{\exp\Big(\tau \, \text{sim}(\mathbf{q}_c, \mathbf{c}_{c,k}^{(d)})\Big)}{\sum_{k=1}^{K_c^{(d)}} \exp\Big(\tau \, \text{sim}(\mathbf{q}_c, \mathbf{c}_{c,k}^{(d)})\Big)} \,.
\end{equation}

Where $\tau$ is a learned temperature parameter. The attention weights $a_{c,k}^{(d)}$ reflect how relevant each prototype is to the current query, with higher weights corresponding to prototypes that are more similar to the query. These attention weights are then used to aggregate the prototypes into a representative prototype for class $c$ in domain $d$:

\begin{equation}
\mathbf{r}_c^{(d)} = \sum_{k=1}^{K_c^{(d)}} a_{c,k}^{(d)} \, \mathbf{c}_{c,k}^{(d)} \,.
\end{equation}

\begin{table*}[htbp]
	\centering
	\caption{Performance comparison of traditional, deep learning, and domain generalization baselines (in \%).}
	\label{tab:dl_comparison}
	\tablebodyfont
	\begin{adjustbox}{width=\textwidth}
		\begin{tabular}{lcccccccc}
			\toprule
			\multirow{2}{*}{Model} & \multicolumn{4}{c}{OSDG1 (\%)} & \multicolumn{4}{c}{OSDG2 (\%)} \\
			\cmidrule(lr){2-5} \cmidrule(lr){6-9}
			& Accuracy & Precision & Recall & F1 & Accuracy & Precision & Recall & F1 \\
			\midrule
			Xgboost\citep{chen2016xgboost} & $26.58 \pm 16.89$ & $19.70 \pm 10.87$ & $33.52 \pm 6.60$ & $20.76 \pm 10.42$ & $56.57 \pm 47.65$ & $68.51 \pm 37.90$ & $57.93 \pm 48.07$ & $57.72 \pm 45.78$ \\      
			SVM\citep{qu2010svm} & $26.02 \pm 13.95$ & $21.45 \pm 19.11$ & $34.34 \pm 3.12$ & $23.62 \pm 15.58$ & $53.74 \pm 47.00$ & $58.96 \pm 48.43$ & $55.30 \pm 47.73$ & $53.25 \pm 46.45$ \\
			KNN\citep{jia2019k} & $37.48 \pm 13.83$ & $42.54 \pm 11.15$ & $34.92 \pm 10.38$ & $29.68 \pm 9.92$ & $60.26 \pm 11.89$ & $64.50 \pm 17.84$ & $63.64 \pm 9.45$ & $56.52 \pm 13.83$ \\
			1DCNN\citep{wu2019one} & $34.64 \pm 13.53$ & $21.56 \pm 1.19$ & $35.20 \pm 6.59$ & $22.64 \pm 6.77$ & $57.38 \pm 16.79$ & $41.21 \pm 24.31$ & $51.77 \pm 15.90$ & $44.09 \pm 22.38$ \\
			2DCNN\citep{xu2018pattern} & $42.15 \pm 11.10$ & $50.94 \pm 2.13$ & $45.42 \pm 6.12$ & $38.32 \pm 8.07$ & $57.31 \pm 15.98$ & $68.52 \pm 19.53$ & $58.39 \pm 20.19$ & $49.44 \pm 24.08$ \\
			LSTM\citep{sherstinsky2020fundamentals} & $23.19 \pm 11.24$ & $15.47 \pm 1.65$ & $30.57 \pm 0.52$ & $18.33 \pm 1.03$ & $67.72 \pm 8.27$ & $57.36 \pm 15.68$ & $62.69 \pm 9.92$ & $57.26 \pm 12.27$ \\ 
			Transformer\citep{vaswani2017attention} & $22.03 \pm 3.55$ & $22.37 \pm 3.39$ & $27.93 \pm 0.81$ & $21.41 \pm 3.95$ & $68.69 \pm 15.87$ & $70.67 \pm 14.38$ & $67.41 \pm 13.47$ & $65.92 \pm 15.19$ \\
			GroupDRO\citep{sagawa2019distributionally} & $57.82 \pm 16.10$ & $58.87 \pm 16.49$ & $52.90 \pm 3.12$ & $48.22 \pm 10.46$ & $71.72 \pm 14.23$ & $75.13 \pm 22.09$ & $74.78 \pm 12.36$ & $68.15 \pm 18.23$ \\
			Coral\citep{sun2016deep} & $52.21 \pm 11.53$ & $57.69 \pm 6.97$ & $49.81 \pm 3.30$ & $44.32 \pm 13.16$ & $70.10 \pm 12.25$ & $85.65 \pm 3.69$ & $71.07 \pm 6.89$ & $64.43 \pm 12.67$ \\
			\bottomrule
		\end{tabular}
	\end{adjustbox}
\end{table*}

\begin{table*}[htbp]
	\centering
	\caption{Few-shot accuracy comparison on OSDG1 and OSDG2 (in \%).}
	\label{tab:meta_accuracy_comparison}
	\tablebodyfont
	\begin{adjustbox}{width=\textwidth}
		\begin{tabular}{lcccccccc}
			\toprule
			\multirow{2}{*}{Model} & \multicolumn{4}{c}{OSDG1 (Shots, \%)} & \multicolumn{4}{c}{OSDG2 (Shots, \%)} \\
			\cmidrule(lr){2-5} \cmidrule(lr){6-9}
			& 1-shot & 3-shot & 5-shot & 10-shot & 1-shot & 3-shot & 5-shot & 10-shot \\
			\midrule
			ProtoNet\citep{snell2017prototypical} & $32.71 \pm 4.93$ & $31.51 \pm 5.57$ & $31.55 \pm 5.84$ & $30.80 \pm 6.88$ & $65.09 \pm 11.65$ & $66.65 \pm 11.55$ & $66.49 \pm 10.77$ & $66.67 \pm 10.22$ \\  
			MAML\citep{finn2017model} & $38.89 \pm 11.39$ & $39.76 \pm 12.63$ & $39.78 \pm 12.13$ & $39.66 \pm 11.87$ & $62.56 \pm 15.59$ & $63.86 \pm 13.69$ & $64.64 \pm 13.73$ & $64.87 \pm 13.86$ \\
			ANIL\citep{raghu2019rapid} & $30.95 \pm 7.54$ & $35.15 \pm 11.99$ & $35.34 \pm 11.87$ & $36.31 \pm 13.42$ & $58.87 \pm 21.49$ & $62.61 \pm 18.63$ & $63.31 \pm 19.08$ & $63.67 \pm 20.21$ \\
			R2D2\citep{bertinetto2018meta} & $41.34 \pm 8.81$ & $40.49 \pm 8.09$ & $40.24 \pm 8.04$ & $40.18 \pm 7.88$ & $54.26 \pm 20.19$ & $63.34 \pm 22.98$ & $64.00 \pm 23.91$ & $64.25 \pm 24.74$ \\
			FEAT\citep{ye2020few} & $47.29 \pm 6.00$ & $49.11 \pm 9.82$ & $49.53 \pm 9.07$ & $48.80 \pm 9.69$ & $53.32 \pm 5.80$ & $57.43 \pm 11.49$ & $57.74 \pm 12.90$ & $58.35 \pm 14.69$ \\
			\textbf{DUPLE (Ours)} & $58.73 \pm 2.39$ & $62.76 \pm 2.65$ & $63.07 \pm 3.56$ & $63.66 \pm 3.23$ & $68.17 \pm 15.34$ & $77.06 \pm 18.98$ & $77.45 \pm 20.35$ & $77.20 \pm 22.73$ \\
			\bottomrule
		\end{tabular}
	\end{adjustbox}
\end{table*}

These representative prototypes $\mathbf{r}_c^{(t)}$ and $\mathbf{r}_c^{(f)}$ are then passed through a cross-domain relation network, which models the relationship between the two domains. The relation vector $\mathbf{v}_c$ captures the consistency between the time-domain and frequency-domain representations for class $c$:
\begin{equation}
\mathbf{v}_c = f_{\text{relation}}([\mathbf{r}_c^{(t)}; \mathbf{r}_c^{(f)}]) \,.
\end{equation}

The relation vector $\mathbf{v}_c$ is passed through a decision network that combines it with the domain importance weights $\alpha_t$ and $\alpha_f$ to produce the final logit for class $c$:
\begin{equation}
\ell_c^{\text{final}} = \alpha_t \, \ell_c^{(t)} + \alpha_f \, \ell_c^{(f)} + h(\mathbf{v}_c, \alpha_t, \alpha_f, \kappa_t, \kappa_f) \,.
\end{equation}

Where $h(\cdot)$ is a fully-connected decision network that integrates the information from both domains. This final logit is used to make the prediction for the query sample, completing the decision process.

\section{Experiments and Analysis}
\label{sec:experiments}

\subsection{Experimental Setup}

All experiments were implemented using PyTorch on an NVIDIA vGPU-32GB platform. We evaluate our framework on two real-world cross-deployment datasets, OSDG1 and OSDG2, utilizing a rigorous Leave-One-Deployment-Out (LODO) cross-validation protocol. Specifically, we iteratively designate one deployment as the unseen target domain (TEST) while training on the remaining deployments (TRAIN), reporting the mean Accuracy, Precision, Recall, and F1 Score across all folds (2 folds for OSDG1, 3 folds for OSDG2). This protocol strictly simulates real-world deployment shifts, where the high variance in our reported results reflects the varying degrees of domain discrepancy across different physical environments.

For meta-learning evaluations, we adopt a standard episodic protocol ($N$-way $K$-shot $Q$-query). We define $N{=}4$ for OSDG1 and $N{=}3$ for OSDG2, with a fixed query size of $Q{=}12$ to ensure class balance. The main comparative experiments are conducted under a 5-shot setting ($K{=}5$), evaluated over 1,000 randomly sampled episodes to ensure statistical reliability. During evaluation, the labeled support samples from the held-out target deployment are used only to construct class prototypes and support-set evidence, while no query labels are used. The model parameters are not updated on the target deployment, ensuring that the evaluation reflects label-scarce cross-deployment recognition rather than target-domain retraining.

\subsection{Cross-Deployment Recognition Performance}
This experiment evaluates whether DUPLE can recognize events in unseen deployment domains rather than only improving performance under random-split recognition. We compare DUPLE with traditional machine learning algorithms, supervised deep learning models, domain generalization (DG) techniques, and representative meta-learning frameworks on OSDG1 and OSDG2.

\begin{table}[htbp]
	\centering
	\caption{Per-deployment macro-F1 breakdown under LODO evaluation. Meta-learning models are evaluated at 5-shot setting.}
	\label{tab:per_deploy_f1_5shot_compact}
	\tablebodyfont
	\begin{adjustbox}{width=\linewidth}
		\begin{tabular}{lccccc}
			\toprule
			\multirow{2}{*}{Model} & \multicolumn{2}{c}{OSDG1 (Deployments, \%)} & \multicolumn{3}{c}{OSDG2 (Deployments, \%)} \\
			\cmidrule(lr){2-3} \cmidrule(lr){4-6}
			& Wire Mesh Fence & Barbed Wire & Hangnail Net & Railing & Wall \\
			\midrule
			GroupDRO & $55.62$ & $40.83$ & $89.05$ & $59.84$ & $55.56$ \\
			ProtoNet & $22.73$ & $34.44$ & $70.98$ & $69.86$ & $54.77$ \\
			FEAT & $39.52$ & $52.68$ & $46.37$ & $67.15$ & $37.98$ \\
			\textbf{DUPLE (Ours)} & $60.48$ & $65.71$ & $95.02$ & $81.80$ & $51.81$ \\
			\bottomrule
		\end{tabular}
	\end{adjustbox}
\end{table}

\begin{table*}[t]
	\centering
	\caption{Few-shot F1 comparison on OSDG1 and OSDG2 (in \%).}
	\label{tab:meta_f1_comparison}
	\tablebodyfont
	\begin{adjustbox}{width=\textwidth}
		\begin{tabular}{lcccccccc}
			\toprule
			\multirow{2}{*}{Model} & \multicolumn{4}{c}{OSDG1 (Shots, \%)} & \multicolumn{4}{c}{OSDG2 (Shots, \%)} \\
			\cmidrule(lr){2-5} \cmidrule(lr){6-9}
			& 1-shot & 3-shot & 5-shot & 10-shot & 1-shot & 3-shot & 5-shot & 10-shot \\
			\midrule
			ProtoNet\citep{snell2017prototypical} & $26.20 \pm 1.40$ & $24.30 \pm 2.08$ & $24.28 \pm 2.19$ & $23.54 \pm 3.08$ & $64.14 \pm 9.90$ & $65.56 \pm 9.83$ & $65.20 \pm 9.05$ & $65.30 \pm 8.52$ \\      
			MAML\citep{finn2017model} & $35.83 \pm 10.87$ & $35.09 \pm 12.51$ & $34.65 \pm 12.43$ & $34.27 \pm 12.12$ & $61.55 \pm 15.00$ & $62.33 \pm 13.11$ & $63.05 \pm 13.21$ & $63.12 \pm 13.49$ \\
			ANIL\citep{raghu2019rapid} & $30.15 \pm 6.99$ & $33.42 \pm 10.34$ & $33.43 \pm 10.19$ & $33.88 \pm 11.25$ & $57.92 \pm 21.77$ & $60.12 \pm 21.38$ & $60.23 \pm 22.78$ & $60.18 \pm 24.63$ \\
			R2D2\citep{bertinetto2018meta} & $38.20 \pm 16.70$ & $37.21 \pm 16.03$ & $36.87 \pm 16.00$ & $36.86 \pm 15.85$ & $54.91 \pm 19.59$ & $63.67 \pm 22.69$ & $64.10 \pm 23.90$ & $64.02 \pm 25.15$ \\     
			FEAT\citep{ye2020few} & $44.28 \pm 7.75$ & $45.43 \pm 10.92$ & $46.10 \pm 9.31$ & $44.30 \pm 9.81$ & $48.71 \pm 7.62$ & $51.25 \pm 13.80$ & $50.50 \pm 15.02$ & $49.97 \pm 16.65$ \\
			\textbf{DUPLE (Ours)} & $58.58 \pm 2.60$ & $62.70 \pm 2.98$ & $63.09 \pm 3.69$ & $63.92 \pm 3.19$ & $66.43 \pm 18.26$ & $75.36 \pm 21.59$ & $76.21 \pm 22.14$ & $76.02 \pm 24.42$ \\
			\bottomrule
		\end{tabular}
	\end{adjustbox}
\end{table*}

\tabref{tab:dl_comparison} summarizes the performance of non-meta-learning approaches. First, traditional machine learning methods such as XGBoost and SVM exhibit limited efficacy in these cross-deployment scenarios. On the OSDG1 dataset, their accuracies remain below 30\%, indicating that manual feature engineering is insufficient to capture invariant patterns amidst severe domain shifts.

Second, standard deep learning models demonstrate significant performance variability driven by architecture bias. On OSDG1, the 2DCNN utilizes Time-Frequency (STFT) representations to achieve 42.15\% accuracy, which is notably superior to raw-signal-based models like the 1DCNN at 34.64\% and the Transformer at only 22.03\%. Conversely, the performance trend on OSDG2 shifts dramatically to favor sequence-based models, where the Transformer and LSTM models achieve leading accuracy rates of 68.69\% and 67.72\%, respectively, significantly outperforming convolutional architectures. This inconsistency highlights that standard supervised learning architectures often overfit to domain-specific feature modalities.

\begin{table*}
	\centering
	\begin{threeparttable}
	\caption{Statistical significance (Welch's $t$-test) against the best meta-learning baseline over 1,000 episodes.}
	\label{tab:significance}
	\tablebodyfont
	\begin{tabular}{lcccc}
		\toprule
		Dataset & Best Baseline $\dagger$ & DUPLE (Ours) & $p$-value & Significant? \\
		\midrule
		OSDG1 & $49.53 \pm 9.07$ (FEAT) & \textbf{$63.07 \pm 3.56$} & $< 10^{-4}$ & \textbf{Yes} \\
		OSDG2 & $66.49 \pm 10.77$ (ProtoNet) & \textbf{$77.45 \pm 20.35$} & $< 10^{-4}$ & \textbf{Yes} \\
		\bottomrule
		\multicolumn{5}{l}{\footnotesize $\dagger$ The best baseline varies by dataset: FEAT for OSDG1, ProtoNet for OSDG2.} \\
	\end{tabular}
	\end{threeparttable}
\end{table*}

Most notably, the Domain Generalization (DG) baselines, GroupDRO and Deep CORAL, demonstrate superior robustness compared to standard Empirical Risk Minimization (ERM) models. To ensure a fair comparison, we implemented these DG algorithms using the same 2DCNN backbone as the standard baseline. By explicitly optimizing for worst-case groups or aligning feature distributions, these methods establish a strong performance baseline. On the challenging OSDG1 dataset, GroupDRO reaches an accuracy of 57.82\%, surpassing the standard 2DCNN by a margin of over 15\%. Similarly, on OSDG2, GroupDRO and Coral maintain the lead with accuracies of 71.72\% and 70.10\%, respectively. However, despite their higher average accuracy, these DG methods exhibit high variance, exemplified by the standard deviation of 16.10\% observed for GroupDRO on OSDG1. This instability underscores a critical limitation: purely invariant representation learning without test-time adaptation is often insufficient to guarantee consistent performance across highly diverse optical fiber deployment environments.

\tabref{tab:meta_accuracy_comparison} and \tabref{tab:meta_f1_comparison} present the few-shot classification results. In general, meta-learning baselines exhibit improved generalization capabilities compared to standard deep models, particularly in low-data regimes.
On OSDG1, FEAT emerges as the strongest baseline with a 5-shot accuracy of $49.53\%$. In contrast, standard metric-based methods such as ProtoNet perform suboptimally, yielding only $31.55\%$. This underperformance suggests that a unimodal single-prototype representation is inadequate for modeling the complex, multi-modal distribution of fiber signals in this highly shifted scenario.
On OSDG2, ProtoNet demonstrates robust performance by achieving $66.49\%$ accuracy, which outperforms more complex architectures like R2D2 and ANIL that reach $64.00\%$ and $63.31\%$ respectively. This indicates that for certain deployment configurations where the domain gap is moderate, the feature space is sufficiently discriminative for Euclidean-distance-based metrics.

The proposed DUPLE framework consistently establishes a new state-of-the-art across both datasets and all shot settings, effectively mitigating the performance gaps observed in baseline methods. Specifically, on the challenging OSDG1 dataset (5-shot), DUPLE achieves an accuracy of 63.07\%, surpassing the best-performing meta-learning baseline, FEAT, by a substantial margin of over 13.5\%. This significant improvement validates the efficacy of our dual-domain fusion strategy in extracting domain-invariant representations under conditions where single-domain baselines struggle. Similarly, in the OSDG2 scenario, our method attains an accuracy of 77.45\%. This performance not only surpasses the strongest meta-learning baseline, ProtoNet, by approximately 11\%, but also outperforms the leading domain generalization baseline, GroupDRO (71.72\%), by a clear margin of 5.73\%.

Notably, we report results as mean $\pm$ standard deviation over LODO folds (OSDG1: 2 deployments; OSDG2: 3 deployments), where the standard deviation reflects cross-deployment heterogeneity rather than random instability. In particular, OSDG2 exhibits larger variability because different held-out deployments pose substantially different difficulty levels. As evidenced by the per-deployment macro-F1 breakdown in \tabref{tab:per_deploy_f1_5shot_compact}, DUPLE achieves $95.02\%$ on Hangnail net and $81.80\%$ on Railing, but drops to $51.81\%$ on Wall, indicating that Wall constitutes a significantly more challenging deployment scenario. Similar cross-deployment fluctuations are observed for ProtoNet and GroupDRO, confirming that the higher standard deviations primarily arise from real deployment shifts captured by the LODO protocol. Overall, DUPLE's superior mean performance amid such heterogeneity confirms its robustness under both moderate and severe domain shifts.

To verify that these performance gains are statistically reliable rather than artifacts of random episodic sampling, we conducted Welch's $t$-tests over 1,000 testing episodes. As detailed in \tabref{tab:significance}, we compared DUPLE against the strongest baseline for each dataset. The resulting $p$-values are significantly below $10^{-4}$ in both scenarios, providing strong statistical evidence that DUPLE yields a significant performance improvement over existing state-of-the-art methods.

In conclusion, DUPLE not only surpasses conventional deep models and DG algorithms in generalization capability but also addresses the representational limitations of existing meta-learners. By effectively fusing dual-domain features with adaptive prototype aggregation, it achieves the highest Accuracy and F1 Scores across heterogeneous deployment environments.

\subsection{Ablation Study}
To systematically dissect the contribution of each component within the DUPLE framework, we conducted ablation studies using the same rigorous Leave-One-Deployment-Out protocol as the main experiments. We evaluated three core modules: the Dual-Domain Multi-Prototype Learner (FPM), the Statistical Guidance Network (SGN), and the Query-Aware Collaborative Decision Making (CDM). The results are summarized in \tabref{tab:ablation_check}.

\begin{table*}[t]
	\centering
	\caption{Ablation study of DUPLE modules at 5-shot setting (in \%).}
	\label{tab:ablation_check}
	\tablebodyfont
	\begin{adjustbox}{width=\textwidth}
		\begin{tabular}{ccc|cccc|cccc}
			\toprule
			\multicolumn{3}{c|}{Module} & \multicolumn{4}{c|}{OSDG1 (\%)} & \multicolumn{4}{c}{OSDG2 (\%)} \\
			\midrule
			FPM & SGN & CDM & Accuracy & Precision & Recall & F1 & Accuracy & Precision & Recall & F1 \\
			\midrule
			-- & -- & -- & $49.86 \pm 3.17$ & $50.22 \pm 2.85$ & $49.86 \pm 3.17$ & $49.28 \pm 3.86$ & $60.03 \pm 20.47$ & $58.98 \pm 22.07$ & $60.03 \pm 20.47$ & $57.35 \pm 22.15$ \\
			\checkmark & -- & -- & $57.79 \pm 2.80$ & $59.09 \pm 4.01$ & $57.79 \pm 2.80$ & $56.20 \pm 1.91$ & $66.62 \pm 24.60$ & $70.81 \pm 21.93$ & $66.62 \pm 24.60$ & $64.89 \pm 26.18$ \\
			\checkmark & \checkmark & -- & $61.07 \pm 0.93$ & $62.33 \pm 2.47$ & $61.07 \pm 0.93$ & $59.88 \pm 0.30$ & $66.70 \pm 24.55$ & $70.94 \pm 21.63$ & $66.70 \pm 24.55$ & $64.98 \pm 26.14$ \\
			\checkmark & -- & \checkmark & $60.14 \pm 0.75$ & $60.85 \pm 0.09$ & $60.14 \pm 0.75$ & $59.20 \pm 1.75$ & $67.53 \pm 23.70$ & $71.46 \pm 20.93$ & $67.53 \pm 23.70$ & $65.67 \pm 25.45$ \\
			\checkmark & \checkmark & \checkmark & \textbf{$63.07 \pm 3.56$} & \textbf{$64.08 \pm 4.35$} & \textbf{$63.07 \pm 3.56$} & \textbf{$63.92 \pm 3.19$} & \textbf{$77.45 \pm 20.35$} & \textbf{$82.87 \pm 11.94$} & \textbf{$77.45 \pm 20.35$} & \textbf{$76.21 \pm 22.14$} \\
			\bottomrule
		\end{tabular}
	\end{adjustbox}
\end{table*}

We conducted systematic ablation experiments to isolate the contributions of the FPM, SGN, and CDM modules, yielding consistent improvements across both datasets. First, the Dual-Domain Multi-Prototype Learner (FPM) acts as the primary performance driver, effectively characterizing intra-class diversity across deployments; it alone boosts accuracy from the baseline's $49.86\%$ to $57.79\%$ on OSDG1. Building on this foundation, the Statistically Guided Network (SGN) and Collaborative Decision Making (CDM) modules provide complementary gains. SGN utilizes sample statistics to regularize domain importance, while CDM dynamically refines prototype weights based on the query. Contrary to unguided aggregation, the inclusion of either module yields distinct performance boosts, confirming that both statistical priors and query-aware attention are essential for robust feature alignment under domain shifts.

Overall, the integration of all three modules in the complete DUPLE framework delivers remarkable results. The full model achieves double-digit accuracy improvements compared to the naive baseline, advancing from $49.86\%$ to $63.09\%$ on OSDG1 and from $60.03\%$ to $77.45\%$ on OSDG2. Notably, while the standard deviation on OSDG2 remains high ($\pm 20.35\%$) due to the intrinsic difficulty variances captured by our Leave-One-Deployment-Out protocol, the mean performance consistently surpasses all ablated variants. By combining dual-domain prototypes with statistical and query-based guidance, our approach significantly mitigates the domain shift problem, offering a stable and accurate solution for diverse deployment environments.

\subsection{Few-Shot Recognition under Label-Scarce Target Deployments}

\begin{figure*}[htbp]
	\centering
	\includegraphics[width=0.98\linewidth]{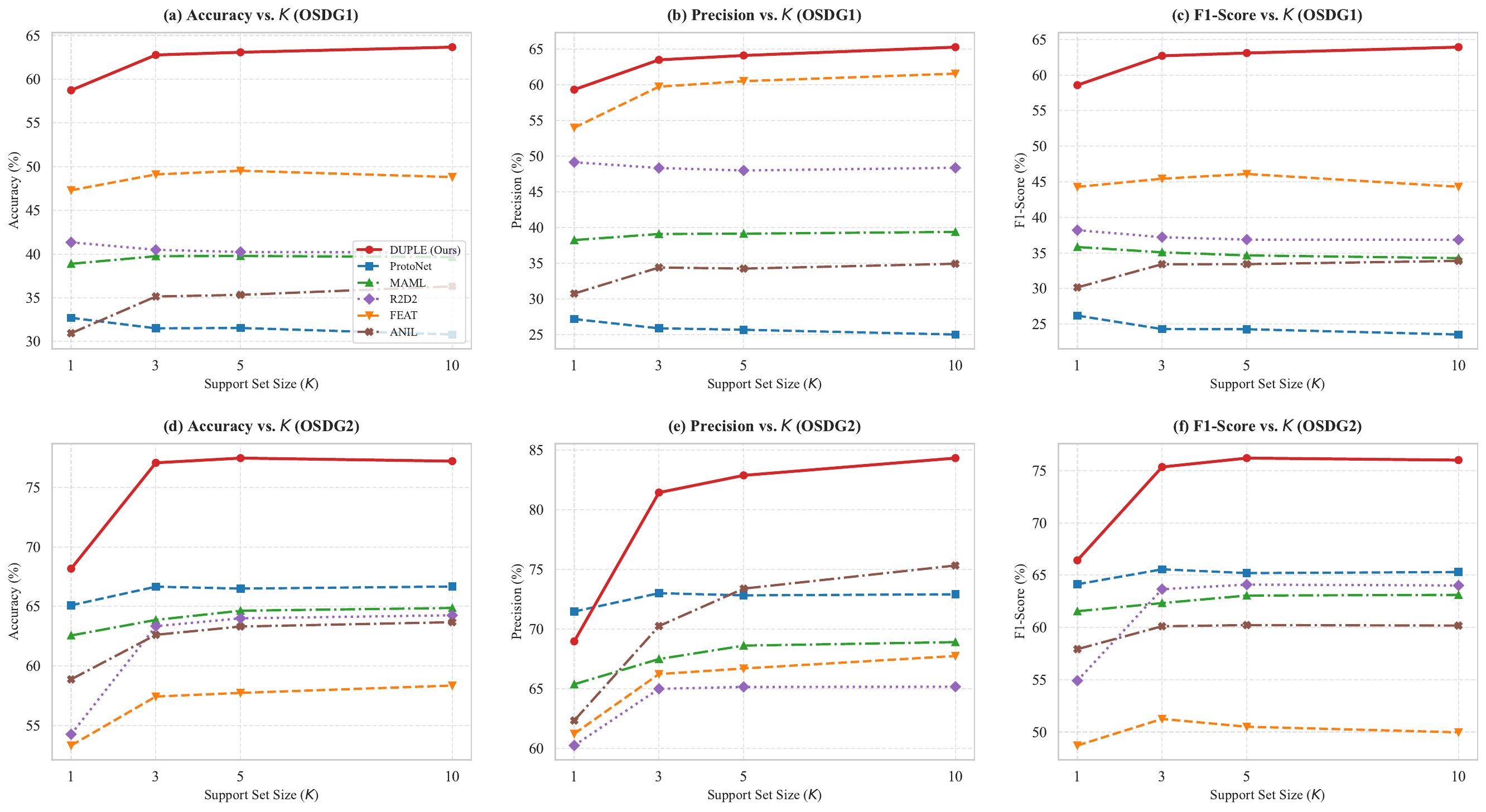}
	\caption{Sensitivity analysis of model performance with respect to support set size $K \in \{1, 3, 5, 10\}$. 
		\textbf{Top Row (OSDG1):} (a-c) DUPLE consistently achieves the highest Accuracy, Precision, and F1-Score, demonstrating exceptional robustness even in the extreme low-data regime ($K=1$), while baselines like ProtoNet struggle significantly. 
		\textbf{Bottom Row (OSDG2):} (d-f) DUPLE exhibits superior data efficiency, showing a rapid performance surge from $K=1$ to $K=3$ and establishing a comprehensive lead in all metrics, unlike baselines that plateau at lower performance levels.}
	\label{fig:sensitivity_all}
\end{figure*}

This analysis simulates early site commissioning, where only a few labeled samples can be collected from a new target deployment. We evaluate support set sizes ranging from $K=1$ to $10$ on both OSDG1 and OSDG2. Beyond Accuracy, we also analyze Precision and F1-Score to assess the trade-off between false alarms and missed detections. \figref{fig:sensitivity_all} presents the results.

On the challenging OSDG1 dataset, DUPLE demonstrates a commanding advantage over all baselines. As illustrated in \figref{fig:sensitivity_all}(a) and (c), our method maintains a significant lead in both Accuracy and F1 Score across all $K$ values. Notably, under extreme single-sample conditions ($K=1$), DUPLE achieves an F1 Score exceeding $58\%$, whereas the strongest baseline (FEAT) hovers around $44\%$, and standard metric-based methods like ProtoNet drop below $27\%$. This substantial gap confirms that the SGN module successfully prevents model collapse even with minimal supervision. Furthermore, as depicted in \figref{fig:sensitivity_all}(b), DUPLE achieves the highest Precision across the board, indicating its ability to learn robust discriminative features without succumbing to the low-precision pitfalls seen in baseline methods.

For the OSDG2 dataset, the results reveal DUPLE's superior learning potential. As shown in \figref{fig:sensitivity_all}(d) and (f), while DUPLE starts with a moderate lead at $K=1$, it exhibits a rapid performance gain as the support set size increases to $K=3$, establishing a clear dominance thereafter.
Crucially, unlike many baselines that trade Recall for Precision, DUPLE achieves the highest performance in both metrics simultaneously. In \figref{fig:sensitivity_all}(e), DUPLE's Precision climbs from $69\%$ at 1-shot to over $84\%$ at 10-shot, significantly outperforming the next best method. This overall advantage is reflected in the F1 score, as shown in \figref{fig:sensitivity_all}(f), where DUPLE's F1 score is more than 10\% higher than that of the best baseline model, ProtoNet. Since the F1 Score represents the harmonic mean of Precision and Recall, DUPLE's top-tier performance on this metric indicates that it provides the most robust solution for practical security monitoring, achieving an optimal balance between false positives and false negatives.

\subsection{Feature Visualization and Efficiency Analysis}

\subsubsection{Feature Space Visualization}

\begin{figure*}[htbp]
	\centering
	
	\begin{subfigure}[b]{0.48\linewidth}
		\centering
		\includegraphics[width=\linewidth]{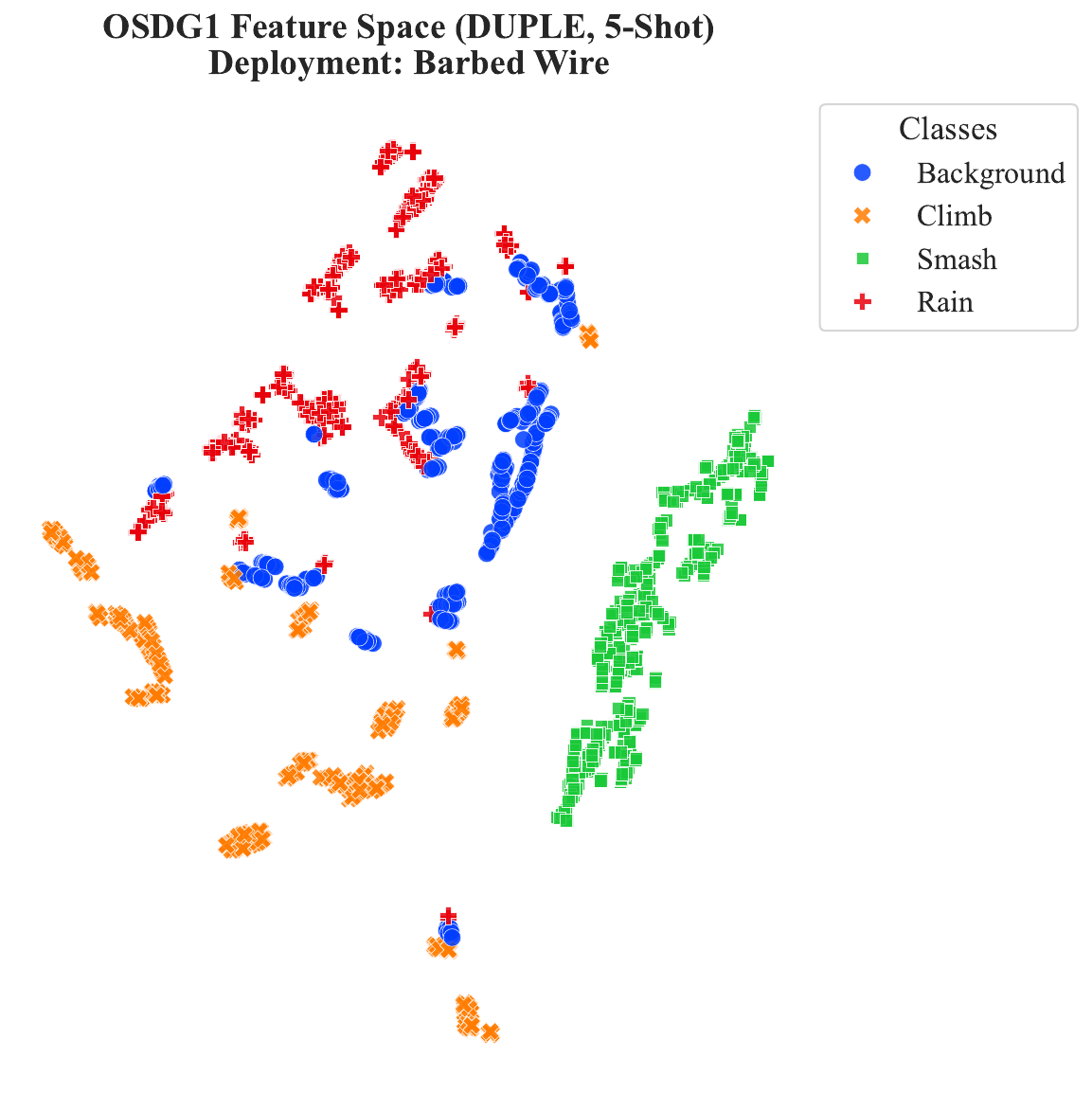}
		\caption{OSDG1 ($K=5$)}
		\label{fig:tsne_osdg1}
	\end{subfigure}
	\hfill % 鍦ㄤ袱鍥句箣闂存拺寮€璺濈
	\begin{subfigure}[b]{0.48\linewidth}
		\centering
		\includegraphics[width=\linewidth]{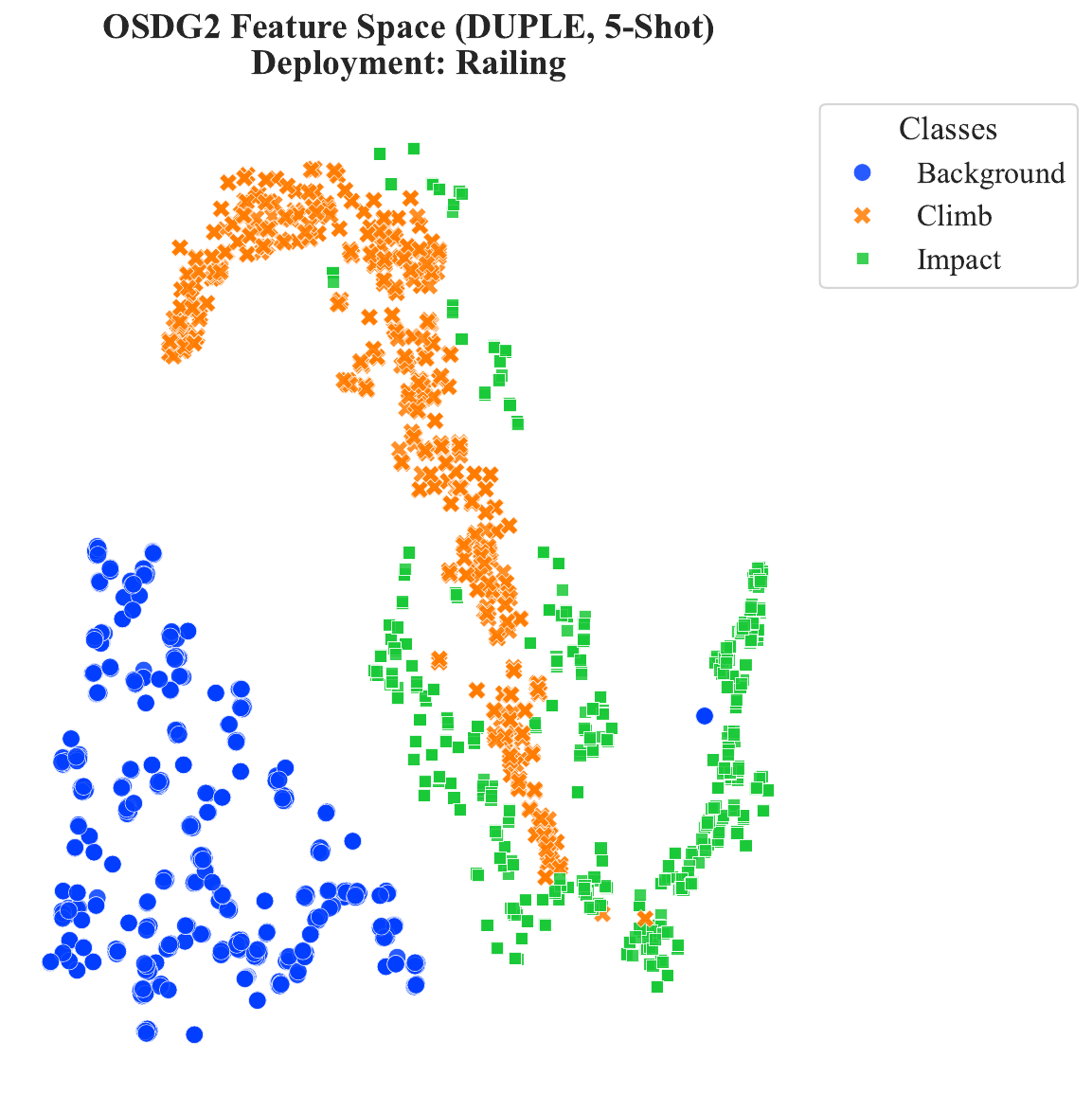}
		\caption{OSDG2 ($K=5$)}
		\label{fig:tsne_osdg2}
	\end{subfigure}
	
	\caption{t-SNE visualization of the feature embeddings learned by DUPLE (5-shot) on representative held-out deployments. 
		\textbf{(a) OSDG1(Barbed wire):} Smash forms a compact and well-separated cluster, whereas Climb is highly fragmented into multiple scattered sub-clusters and shows noticeable mixing with Background and Rain. This pattern reflects strong deployment-induced variability and helps explain the more challenging performance on OSDG1.
		\textbf{(b) OSDG2(Railing):}The embeddings exhibit clearer class separation among Background, Climb, and Impact, with substantially reduced inter-class mixing compared with OSDG1, consistent with the stronger quantitative results on OSDG2.}
	\label{fig:tsne}
\end{figure*}

To qualitatively assess the discriminative structure of the learned representations, we employ t-SNE to project the high-dimensional query embeddings into a two-dimensional space for visualization. Under the rigorous Leave-One-Deployment-Out (LODO) protocol, the feature geometry may vary across folds due to deployment-specific signal characteristics. To provide a representative and reproducible view without cherry-picking, we report the embeddings from the most challenging deployment fold for each dataset, identified by the lowest silhouette score in the learned embedding space. Accordingly, the visualizations correspond to the Barbed wire deployment for OSDG1 and the Railing deployment for OSDG2 (\figref{fig:tsne}).

\begin{table}[htbp]
	\centering
	\caption{Computational efficiency comparison.}
	\label{tab:efficiency}
	\tablebodyfont
	\begin{adjustbox}{width=\columnwidth}
		\begin{tabular}{lcccc}
			\toprule
			Model & Params (M) & FLOPs (G) & Time (ms) & FPS \\
			\midrule
			1D-CNN        & 0.01 & 0.11 & 0.13 & 7690.11 \\
			2D-CNN        & 0.04 & 0.74 & 0.15 & 6840.92 \\
			Transformer   & 1.19 & 0.31 & 0.61 & 1634.76 \\
			ProtoNet      & 2.44 & 1.65 & 1.46 & 685.71 \\
			\textbf{DUPLE (Ours)} & 2.48 & 3.30 & 6.12 & 163.42 \\
			\bottomrule
		\end{tabular}
	\end{adjustbox}
\end{table}

In the more challenging OSDG1 setting, \figref{fig:tsne}(a) reveals a heterogeneous embedding structure under strong deployment-induced variations. The Smash samples form a compact and well-separated cluster, whereas Climb exhibits pronounced fragmentation into multiple scattered sub-clusters, with noticeable mixing with Background and Rain. This pattern reflects substantial intra-class variability and deployment-specific distribution shifts, which is consistent with the more difficult quantitative results reported in \tabref{tab:dl_comparison}. Importantly, despite the fragmented structure, the embeddings do not collapse into a single indistinguishable region, indicating that DUPLE still preserves meaningful class-related structures under this challenging deployment.

In contrast, \figref{fig:tsne}(b) shows that OSDG2 (Railing) yields a more structured organization of the embedding space, with clearer separation among Background, Climb, and Impact and reduced inter-class mixing compared with OSDG1. This observation aligns with the stronger performance of DUPLE on OSDG2, suggesting that the proposed dual-domain and prototype-based metric learning can produce more consistent class-wise clustering under certain deployment conditions.

\subsubsection{Computational Efficiency Analysis}

To assess the deployment feasibility of the proposed method in real-time monitoring systems, we benchmarked the model complexity (Parameters), computational cost (FLOPs), and inference latency against representative baselines. All evaluations were conducted on a single NVIDIA VGPU-32GB using a batch size of 1 to simulate the sequential data processing typical of edge computing environments. The standardized input length was set to 150,000 sampling points.

\tabref{tab:efficiency} summarizes the efficiency metrics. Traditional deep learning models, especially 1D-CNN and 2D-CNN, have the lowest computational overhead, with inference times below 0.2 milliseconds. However, their limited parameter capacities of 0.01 M and 0.04 M, respectively, restrict feature representation and lead to poor generalization performance. While Transformer models maintain a moderate inference speed, their accuracy is lower compared to meta-learning methods, indicating that standard attention mechanisms alone are insufficient for this specific cross-domain few-shot task.

\begin{table*}[htbp]
	\centering
	\caption{Per-class accuracy on OSDG1 and OSDG2.}
	\label{tab:merged_per_class_acc}
	\tablebodyfont
	\begin{adjustbox}{width=\textwidth}
		\begin{tabular}{l|cccc|ccc}
			\toprule
			\multirow{2}{*}{Model} & \multicolumn{4}{c}{OSDG1} & \multicolumn{3}{|c}{OSDG2} \\
			\cmidrule(lr){2-5} \cmidrule(lr){6-8}
			& Background & Climb & Rain & Smash & Background & Climb & Impact \\
			\midrule
			XGBoost & $24.40 \pm 24.40$ & $67.04 \pm 29.21$ & $4.25 \pm 3.44$ & $38.37 \pm 38.37$ & $73.59 \pm 37.35$ & $56.50 \pm 41.17$ & $45.50 \pm 41.32$ \\
			SVM & $9.60 \pm 9.60$ & $75.97 \pm 14.72$ & $43.92 \pm 6.08$ & $7.86 \pm 7.86$ & $63.16 \pm 38.85$ & $64.33 \pm 45.58$ & $39.17 \pm 39.07$ \\
			KNN & $12.11 \pm 4.11$ & $3.72 \pm 1.28$ & $59.24 \pm 8.07$ & $64.60 \pm 34.60$ & $68.83 \pm 31.03$ & $39.06 \pm 34.54$ & $83.03 \pm 17.46$ \\
			1DCNN & $0.00 \pm 0.00$ & $0.00 \pm 0.00$ & $63.94 \pm 36.06$ & $76.87 \pm 17.41$ & $28.67 \pm 40.54$ & $38.99 \pm 37.88$ & $87.67 \pm 10.00$ \\
			2DCNN & $43.10 \pm 38.30$ & $59.46 \pm 40.54$ & $37.51 \pm 30.78$ & $41.63 \pm 11.23$ & $64.32 \pm 41.77$ & $44.69 \pm 39.11$ & $66.16 \pm 46.78$ \\
			LSTM & $27.03 \pm 27.03$ & $32.56 \pm 32.56$ & $46.15 \pm 46.15$ & $16.53 \pm 14.90$ & $54.65 \pm 40.19$ & $43.92 \pm 33.25$ & $89.50 \pm 13.46$ \\
			Transformer & $2.00 \pm 1.20$ & $42.01 \pm 19.24$ & $25.29 \pm 0.29$ & $42.43 \pm 22.43$ & $50.13 \pm 8.57$ & $70.32 \pm 32.78$ & $81.78 \pm 15.63$ \\
			GroupDRO & $18.60 \pm 18.60$ & $58.83 \pm 39.92$ & $70.13 \pm 1.02$ & $64.03 \pm 31.17$ & $98.72 \pm 1.81$ & $58.96 \pm 39.27$ & $66.67 \pm 47.14$ \\
			Coral & $11.63 \pm 11.63$ & $26.20 \pm 22.45$ & $70.66 \pm 28.38$ & $90.77 \pm 3.63$ & $78.52 \pm 28.76$ & $34.68 \pm 45.50$ & $100.00 \pm 0.00$ \\
			\midrule
			ProtoNet & $59.25 \pm 12.36$ & $24.80 \pm 24.80$ & $37.52 \pm 31.43$ & $4.62 \pm 2.47$ & $51.65 \pm 33.60$ & $73.28 \pm 15.32$ & $74.56 \pm 23.49$ \\
			MAML & $65.41 \pm 29.01$ & $61.46 \pm 9.45$ & $3.29 \pm 0.15$ & $28.95 \pm 14.90$ & $56.10 \pm 22.56$ & $59.41 \pm 25.83$ & $78.40 \pm 23.72$ \\
			ANIL & $45.13 \pm 24.69$ & $39.36 \pm 4.13$ & $15.09 \pm 5.73$ & $41.77 \pm 18.73$ & $59.07 \pm 38.61$ & $61.95 \pm 25.35$ & $68.91 \pm 32.57$ \\
			R2D2 & $32.14 \pm 20.66$ & $84.75 \pm 14.83$ & $31.40 \pm 5.66$ & $12.68 \pm 11.23$ & $68.94 \pm 29.44$ & $58.28 \pm 30.93$ & $64.79 \pm 12.28$ \\
			FEAT & $23.03 \pm 2.10$ & $65.06 \pm 34.94$ & $44.81 \pm 26.66$ & $66.49 \pm 30.57$ & $55.59 \pm 32.83$ & $42.45 \pm 42.02$ & $75.17 \pm 35.11$ \\
			\textbf{DUPLE (Ours)} & \textbf{$69.25 \pm 29.14$} & \textbf{$64.87 \pm 1.56$} & \textbf{$52.13 \pm 23.19$} & \textbf{$65.99 \pm 17.63$} & \textbf{$71.60 \pm 36.82$} & \textbf{$86.41 \pm 6.78$} & \textbf{$74.35 \pm 19.37$} \\
			\bottomrule
		\end{tabular}
	\end{adjustbox}
\end{table*}

Compared to the ProtoNet baseline model, our proposed DUPLE framework introduces higher computational load and latency, at 3.30 G FLOPs and 6.12 ms, respectively. This increase is attributed to the parallel processing architecture of the dual-domain branches and the additional computation required for instance-level reweighting by the SGN module. Although DUPLE is slower than lightweight baselines, it still satisfies real-time DFOS monitoring with a large margin (6.12~ms per ${\sim}3$~s window). Nevertheless, DUPLE achieves an inference speed of approximately 163 FPS. Considering that typical distributed fiber optic sensing systems require alarm response frequencies of 20 Hz to 50 Hz, DUPLE's processing speed is sufficient to support real-time operation. Therefore, the proposed method achieves a practical trade-off, significantly improving accuracy without violating latency constraints in industrial deployments.

\subsection{Per-Deployment and Per-Class Analysis}

While overall accuracy provides a general measure of performance, it can obscure critical failures in specific classes, particularly under the imbalance induced by domain shifts. Therefore, we conduct a fine-grained analysis to determine whether models maintain reliable detection capabilities across all intrusion types or merely overfit to majority classes.

Analysis of OSDG1. As detailed in \tabref{tab:merged_per_class_acc}, standard deep learning baselines on OSDG1 suffer from catastrophic mode collapse. Notably, the 1DCNN failed to identify a single sample from the Background or Climb classes, yielding 0.00\% accuracy in both cases, likely due to overfitting on the Smash class where it reached 76.87\%. Even robust Domain Generalization (DG) methods exhibited significant instability; while GroupDRO performed well on Rain and Smash, it struggled severely with the Background class, achieving an accuracy of only 18.60\%. This indicates that optimizing for worst-case groups does not fully resolve feature ambiguity in this domain.

\begin{figure*}[htbp]
	\centering
	\includegraphics[width=0.95\linewidth]{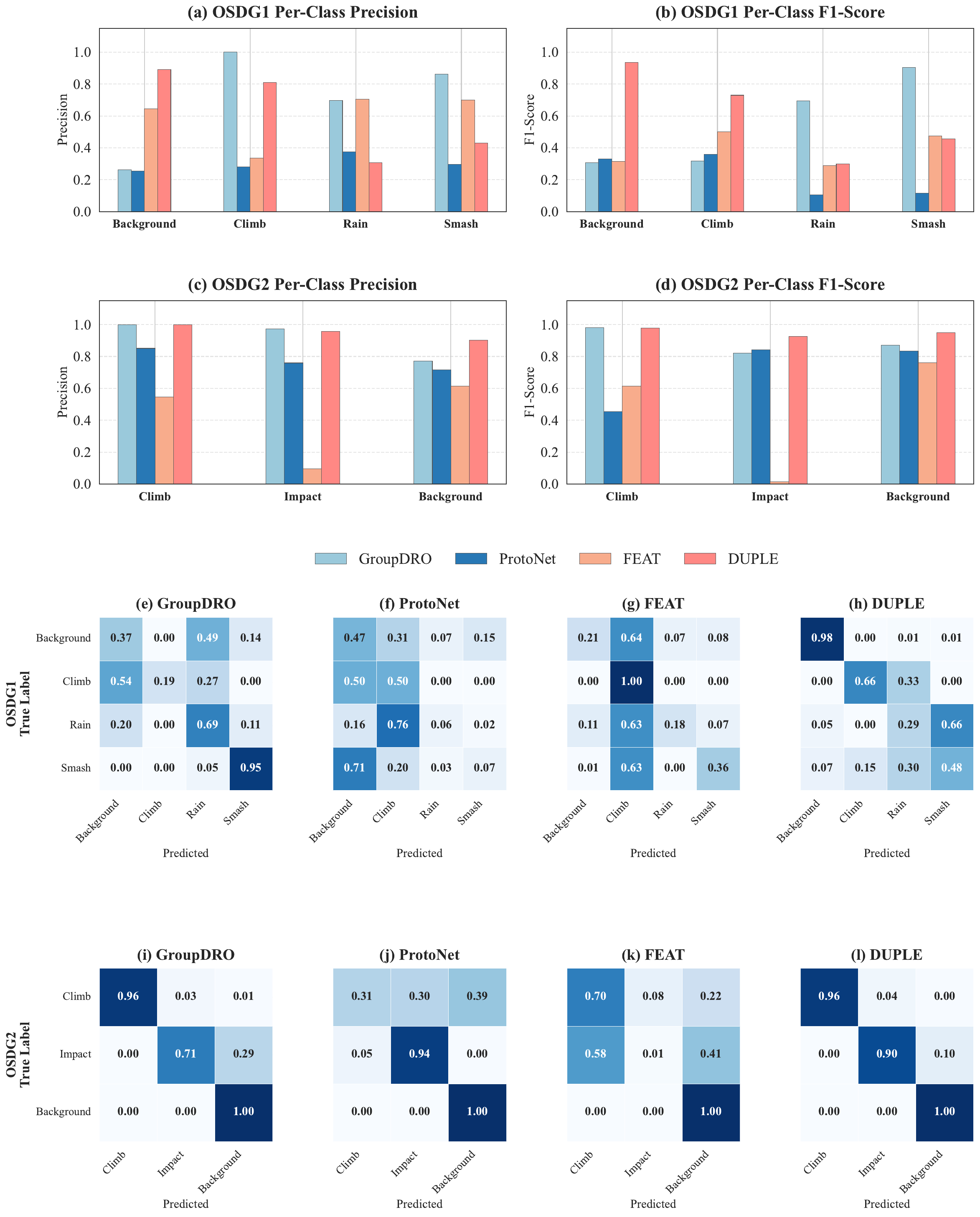}
	\caption{Per-class performance and confusion matrix analysis on OSDG1 and OSDG2.}
	\label{fig:cm_comparison}
\end{figure*}

Meta-learning baselines also demonstrated severe class selectivity. ProtoNet performed adequately on Background samples but its detection rate for Smash events dropped to a negligible 4.62\%. These disjointed decision boundaries are visually corroborated by the confusion matrices in \figref{fig:cm_comparison}. Specifically, GroupDRO (\figref{fig:cm_comparison}e) shows substantial confusion between Background and Rain, while ProtoNet (\figref{fig:cm_comparison}f) fails to establish a clear diagonal for the Smash class.

In contrast, the proposed DUPLE framework successfully mitigated these extreme disparities. DUPLE maintained balanced detection capabilities across all categories. Even on the most challenging Rain class, it retained a viable accuracy of 52.13\%, a marked improvement over the near-zero failures observed in baseline methods. As illustrated in \figref{fig:cm_comparison}(h), DUPLE recovers the diagonal structure of the confusion matrix, securing a consistent performance range of 52\% to 69\% across all four classes. This demonstrates that DUPLE prioritizes global reliability over peak performance on easier categories.

Analysis of OSDG2. Although the domain gap narrows on OSDG2, baseline models continued to exhibit distinct failure modes. Coral achieved perfect detection on Impact samples but sacrificed performance on the Climb class, where accuracy fell to 34.68\%. Similarly, GroupDRO was heavily skewed, identifying 98.72\% of Background samples but correctly classifying only 58.96\% of Climb instances.

Among meta-learners, DUPLE demonstrated superior decision refinement. Quantitatively, DUPLE outperformed both GroupDRO and ProtoNet on the difficult Climb class with a leading accuracy of 86.41\%, while maintaining strong performance on Background at 71.60\% and Impact at 74.35\%. The visual comparison between \figref{fig:cm_comparison}(i) and \figref{fig:cm_comparison}(l) further highlights this advantage: while GroupDRO misclassifies a significant portion of Impact samples as Background, DUPLE reduces this error and exhibits sharper diagonal dominance, indicating more precise decision boundaries.

In summary, the primary failure mode of baseline models across both datasets was single-class collapse caused by domain shift. DUPLE's dual-domain statistically guided prototyping strategy effectively reduces this class-wise disparity, ensuring that performance remains viable even for the worst-case categories.

\section{Conclusion}
\label{sec:conclusion}

This paper proposed DUPLE, an intelligent cross-deployment recognition framework for DFOS-based perimeter security under label-scarce target deployments. DUPLE addresses deployment-induced distribution shifts by combining dual-domain signal modeling, multi-prototype class representation, statistical reliability guidance, and query-aware prototype aggregation. Instead of relying on a single fixed class prototype or a static domain-invariant classifier, DUPLE constructs adaptive class evidence from limited support samples and adjusts the contribution of time-domain and time-frequency information according to sample-specific statistical cues.

Experiments on two real-world DFOS benchmarks, OSDG1 and OSDG2, demonstrate the effectiveness of DUPLE under a Leave-One-Deployment-Out evaluation protocol. Compared with traditional machine learning methods, supervised deep models, domain generalization baselines, and representative meta-learning approaches, DUPLE achieves consistently better recognition performance across unseen deployments. Few-shot experiments further show that DUPLE remains effective when only a small number of labeled samples are available from a new target deployment. In addition, ablation studies, per-deployment analysis, per-class evaluation, feature visualization, and efficiency analysis confirm that the proposed components provide complementary benefits and improve both overall recognition accuracy and class-wise robustness.

Although the present study focuses on interferometric fiber-optic vibration signals and single-event recognition, the results suggest that statistically guided meta-learning is a promising direction for practical fiber-optic perimeter security systems. Future work will extend DUPLE to multi-channel sensing, multi-event recognition, online adaptation, and long-term field monitoring under changing environmental and deployment conditions.
%%       \cite{<label>} ==> [1]

%%
\section*{CRediT Authorship Contribution Statement}

\textbf{Yifan He}: Conceptualization, Methodology, Software, Investigation, 
Data curation, Formal analysis, Visualization, Writing - original draft;\textbf{Haodong Zhang}: Conceptualization, Methodology, Investigation, Writing - review \& editing;\textbf{Qiuheng Song}: Conceptualization, Resources;\textbf{Lin Lei}: Conceptualization, Data curation;\textbf{Zhenxuan Zeng}: Conceptualization, Writing - review \& editing;\textbf{Haoyang He}: Investigation, Data curation;\textbf{Hongyan Wu}: Conceptualization, Methodology, Supervision, Writing - review \& editing.

\section*{Declaration of Competing Interest}

Author Qiuheng Song is a shareholder of Sichuan Fujinan Technology Co., Ltd., which provided the data used
in this study. Sichuan Fujinan Technology Co., Ltd. had no role in the study design, data analysis,
interpretation of the results, or the decision to submit the manuscript for
publication. The other authors declare that they have no competing financial
interests or personal relationships that could have influenced the work
reported in this paper.

\section*{Acknowledgments}

The authors would like to thank Sichuan Fujinan Technology Co., Ltd. for providing the distributed fiber-optic sensing data and for their technical assistance with the experimental deployments.

\section*{Declaration of Generative AI and AI-Assisted Technologies in the Writing Process}

During the preparation of this work, the authors used OpenAI's ChatGPT for language polishing, structural editing, and formatting refinement. The tool was not used to generate research data, perform experiments, or draw scientific conclusions. The authors reviewed and edited the output as needed and take full responsibility for the content of the published article.


%%\bibliographystyle{elsarticle-num}  % 鎴栬€?plain, IEEEtran, etc.

%%\bibliography{refs}
\end{document}